\pdfoutput=1
\documentclass[journal,twocolumn,10pt]{IEEEtran}

\usepackage[T1]{fontenc}
\usepackage[utf8]{inputenc}
\usepackage{amsmath,amssymb,amsfonts,bm}
\usepackage{graphicx}
\usepackage{cite}
\usepackage{url}
\usepackage{array}
\usepackage{tabularx}
\usepackage[table]{xcolor}
\usepackage{booktabs}
\usepackage{placeins}
\usepackage{stfloats}
\usepackage{pifont}
\newif\ifcameraready
\camerareadyfalse
\definecolor{navcol}{RGB}{0,96,112}       
\usepackage{hyperref}
\ifcameraready
  \hypersetup{hidelinks}
\else
  \hypersetup{colorlinks=true,allcolors=navcol,breaklinks=true}
\fi
\hypersetup{pdftitle={Hierarchical Adaptive Feature Refinement Network for VHR Remote Sensing Image Segmentation},
  pdfauthor={Shuaishuai Cao and Meng Tang and Shuwei Peng and Xuan Liu and Min Huang and Jie Chen and Jiacheng Niu and Yong Chen and Edore Akpokodje and Hui Lin},
  pdfkeywords={frequency-domain refinement, hierarchical feature fusion, semantic segmentation, confusion-aware decoding, very-high-resolution remote sensing}}


\definecolor{bestcol}{RGB}{150,15,30}     
\definecolor{seccol}{RGB}{20,80,150}      
\definecolor{rsgain}{RGB}{0,110,74}       
\definecolor{abldown}{RGB}{105,105,105}   
\definecolor{tabstd}{RGB}{110,110,110}    
\definecolor{rsmark}{RGB}{238,244,251}    
\definecolor{rshead}{RGB}{246,246,246}    
\definecolor{rsrule}{RGB}{0,0,0}          
\newcommand{\tabtoprule}{\specialrule{1.0pt}{0pt}{0pt}}
\newcommand{\tabheadrule}{\specialrule{0.5pt}{0pt}{0pt}}
\newcommand{\tabbotrule}{\specialrule{1.0pt}{0pt}{0pt}}
\newcommand{\oursrow}[1]{\textbf{#1}}
\newcommand{\bestv}[1]{{\color{bestcol}\textbf{#1}}}
\newcommand{\secv}[1]{{\color{seccol}\underline{#1}}}
\newcommand{\std}[1]{{\scriptsize\color{tabstd}$\pm$#1}}
\newcommand{\down}[1]{{\scriptsize\color{abldown}$\downarrow$#1}}
\newcolumntype{G}{>{\color{rsgain}}c}
\newcolumntype{L}{>{\raggedright\arraybackslash}X}
\newcolumntype{C}{>{\centering\arraybackslash}X}
\newcommand{\cmark}{\ding{51}}

\newcommand{\tabnote}[1]{%
  \par\addvspace{2.5pt}\noindent\parbox{\linewidth}{\footnotesize #1}\par}
\newcommand{\bestsec}{The best and second-best entries in each column are
  \bestv{bold red} and \secv{underlined blue}, respectively.}

\newcommand{\figref}[1]{Fig.~\ref{fig:#1}}
\newcommand{\tabref}[1]{Table~\ref{tab:#1}}
\newcommand{\secref}[1]{Section~\ref{sec:#1}}
\newcommand{\equref}[1]{\eqref{eq:#1}}

\newcommand{\figrange}[2]{Figs.~\ref{fig:#1}--\ref{fig:#2}}
\newcommand{\tabrange}[2]{Tables~\ref{tab:#1}--\ref{tab:#2}}
\newcommand{\equrange}[2]{\eqref{eq:#1}--\eqref{eq:#2}}
\newcommand{\Hetero}{\mathcal{H}}
\newcommand{\hafr}{HAFR-Net}

\newcommand{\repolink}[1]{\url{#1}}

\makeatletter
\def\section{\@startsection{section}{1}{\z@}{3.0ex plus 0.3ex minus 1.5ex}%
{0.7ex plus 0.15ex minus 0ex}{\normalfont\normalsize\centering\scshape}}
\def\subsection{\@startsection{subsection}{2}{\z@}{3.5ex plus 0.3ex minus 1.5ex}%
{0.7ex plus 0.15ex minus 0ex}{\normalfont\normalsize\itshape}}
\makeatother

\newcommand{\email}[1]{{\urlstyle{same}\nolinkurl{#1}}}

\begin{document}

\title{Hierarchical Adaptive Feature Refinement Network\\
for VHR Remote Sensing Image Segmentation}

\author{Shuaishuai~Cao, Meng~Tang, Shuwei~Peng, Xuan~Liu, Min~Huang, Jie~Chen, Jiacheng~Niu, Yong~Chen, Edore~Akpokodje, and Hui~Lin,~\IEEEmembership{Senior~Member,~IEEE}%
\thanks{This work has been submitted to the IEEE for possible publication.  Copyright may be transferred without notice, after which this version may no longer be accessible.}
}

\pagestyle{plain}

\maketitle
\thispagestyle{plain}

\begin{abstract}
Semantic segmentation of very-high-resolution (VHR) remote sensing imagery increasingly benefits from strong pretrained hierarchical encoders, yet exploiting their multi-stage representations remains difficult.  Nearby regions demand different balances between fine detail and semantic context, aggressive task-specific transformations perturb useful pretrained features, and conventional semantic supervision provides limited structural guidance.  We present \hafr{}, a progressive refinement framework that adaptively organizes and conservatively refines hierarchical representations instead of replacing them with a monolithic decoder transformation.  Heterogeneity-Guided Stage-Adaptive Fusion (HG-SAF) predicts dense stage weights conditioned on local feature variation.  A Frequency-Residual Adapter (FRA) then injects frequency information through a bounded, zero-initialized residual branch that keeps the fused representation as its reference.  A Confusion-Aware Tri-Prior Decoder (CATP) finally regularizes the prediction with boundary, objectness, and training-derived class-relation cues.  Under a matched Swin-B training and single-scale inference protocol, \hafr{} attains 84.12\%, 87.86\%, 55.17\%, and 67.70\% mIoU on ISPRS Vaihingen, ISPRS Potsdam, LoveDA, and OpenEarthMap, improving the matched UPerNet baseline by 0.55, 0.95, 1.55, and 1.84 percentage points.  Controlled analyses further show consistent spatial reweighting beyond content-only routing, improved boundary and thin-structure accuracy over matched spatial and spectral alternatives, and reduced confusion on pre-declared class pairs.  The trained LoveDA model is released, with the source code to follow, at \repolink{https://github.com/anticipate218/HAFRNet}.
\end{abstract}

\begin{IEEEkeywords}
Frequency-domain refinement, hierarchical feature fusion, semantic segmentation, confusion-aware decoding, very-high-resolution (VHR) remote sensing.
\end{IEEEkeywords}


\section{Introduction}
\label{sec:intro}

\IEEEPARstart{V}{ery}-high-resolution (VHR) remote sensing imagery provides detailed observations of buildings, roads, vehicles, vegetation, and other land-cover elements at fine spatial scales~\cite{vhr_survey,rs_review2023,xbd}. Modern segmentation systems based on CNNs, Transformers, and state-space models have benefited substantially from large pretrained encoders and hierarchical feature representations~\cite{long2015fcn,unet,pspnet,deeplabv3plus,swin,segformer,gu2024mamba,pyramidmamba}. As encoder capacity continues to improve, however, a growing part of the segmentation problem shifts from representation extraction to representation utilization: the decoder must determine how to combine spatially detailed shallow features with semantically strong deep features, and how to refine them without unnecessarily destroying information already encoded by pretraining.

\begin{figure*}[t]
\centering
\includegraphics[width=0.92\textwidth]{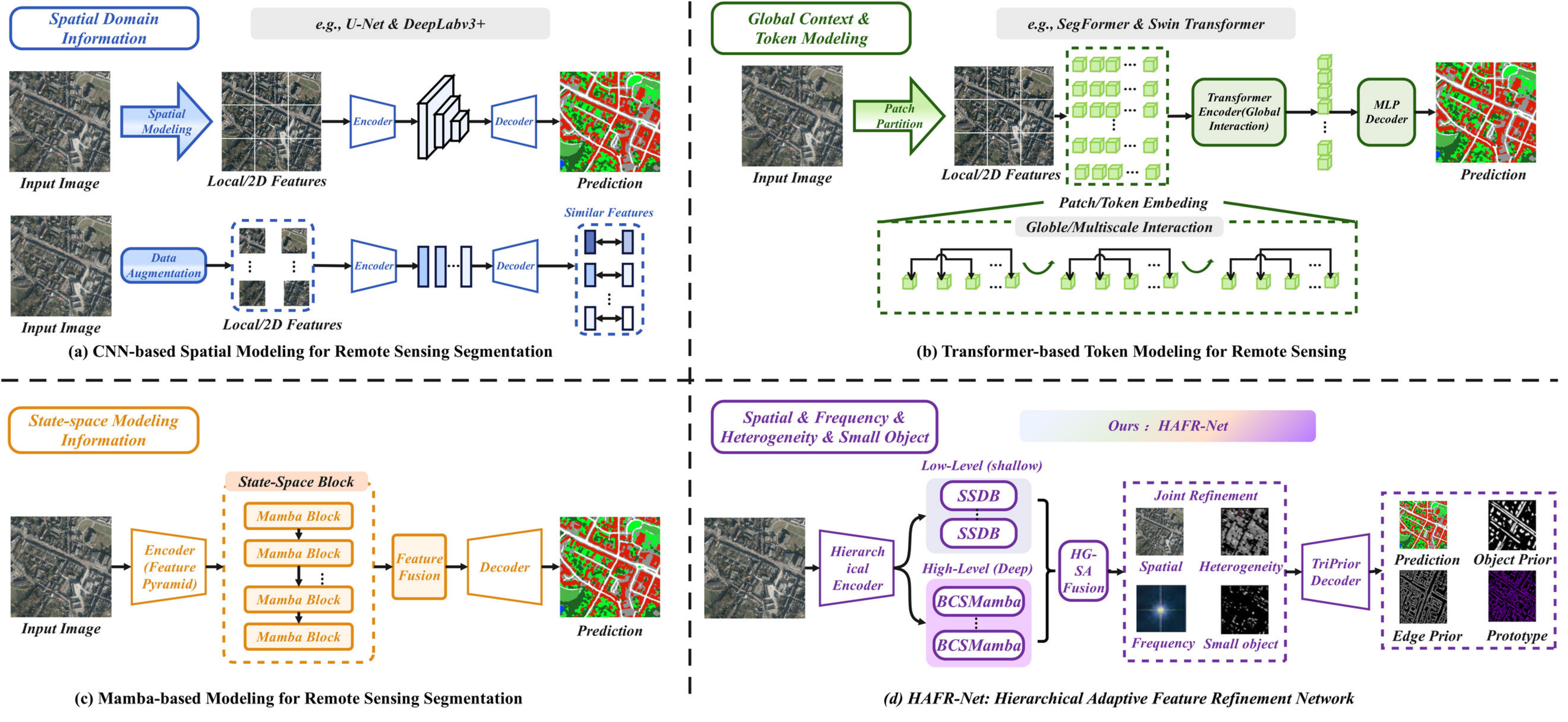}
\caption{Positioning of \hafr{}.  (a)--(c)~CNN, Transformer and state-space decoders combine the encoder hierarchy through a fixed aggregation rule.  (d)~\hafr{} keeps the pretrained hierarchy and refines it progressively, through stage-specialized preparation, heterogeneity-guided stage-adaptive fusion, bounded residual frequency adaptation, and confusion-aware tri-prior decoding.}
\label{fig:teaser}
\end{figure*}

A key difficulty in VHR imagery is that the preferred balance between spatial detail and semantic abstraction varies strongly within the same image (\figref{teaser}). Small vehicles, narrow roads, roof edges, and object boundaries benefit from high-resolution stages, whereas large roofs, agricultural parcels, and homogeneous regions often require deeper contextual representations. Nevertheless, many commonly used decoders aggregate multi-stage features through fixed top-down connections, predefined summation, or image-level weighting~\cite{fpn,xie2018uper,unetformer,dcswin}. Such strategies are effective in general, but they provide limited flexibility when the desired stage contribution changes from pixel to pixel within a highly heterogeneous scene.

A second consideration concerns how additional decoder capacity is introduced. Frequency-domain and global-context operators can enhance detail or long-range interactions~\cite{gfnet,ffc,sffnet,fdnet,afenet}, but unrestricted transformations may require the optimizer to simultaneously preserve pretrained representations and learn a new task-specific mapping. For VHR datasets, which are often considerably smaller than generic pretraining corpora, a more conservative strategy is attractive: rather than replacing the fused representation, the decoder should introduce bounded residual corrections that remain close to a known reference behavior at the beginning of fine-tuning.

Finally, accurate VHR segmentation is not determined by region classification alone. Errors frequently occur around object boundaries, small foreground regions, and semantically similar categories~\cite{gscnn,boundaryloss,wang2022contrastive,protoseg}. These patterns motivate the use of structural supervision and class-relation constraints during decoding. Rather than treating such auxiliary signals as independent tasks, they can be used as lightweight regularizers that shape the final feature space while leaving inference dominated by a single semantic prediction branch. When class-relation pairs are used, they are identified exclusively from a training-only pilot split and are fixed before final evaluation.

Based on these observations, we propose \hafr{}, a Hierarchical Adaptive Feature Refinement Network for VHR remote sensing image segmentation. The framework follows a progressive refinement philosophy. A pretrained Swin-B encoder first provides four hierarchical feature stages. Heterogeneity-Guided Stage-Adaptive Fusion (HG-SAF) adaptively aggregates these features with pixel-wise weights conditioned on local heterogeneity. Frequency-Residual Adapter (FRA) then performs a bounded residual correction in the frequency domain instead of replacing the fused representation with a full spectral transform. Finally, the Confusion-Aware Tri-Prior Decoder (CATP) regularizes semantic decoding using structural and class-relation cues. The resulting decoder explicitly separates adaptive hierarchical fusion, conservative feature refinement, and final tri-prior decoding.

The main contributions are as follows.
\begin{itemize}
    \item We formulate VHR decoding as hierarchical adaptive refinement rather than generic feature aggregation. HG-SAF predicts pixel-wise stage weights from local feature heterogeneity, enabling the contribution of shallow spatial detail and deep semantic context to vary across a scene.
    \item We introduce a bounded residual refinement strategy for fused hierarchical representations. FRA performs channel-bottlenecked spectral modulation with an explicitly bounded gate and residual initialization, refining detail while starting from the pretrained spatial representation rather than an unconstrained frequency transform.
    \item We combine adaptive fusion and conservative refinement with confusion-aware tri-prior decoding and validate each design through matched controls. Experiments on four VHR benchmarks include controlled decoder comparisons, direct spatial and frequency alternatives, stage-weight analysis, boundary metrics, class-confusion analysis, statistical uncertainty, and efficiency measurements.
\end{itemize}

\section{Related Work}
\label{sec:related}

\subsection{Hierarchical Feature Fusion for Remote Sensing Segmentation}
\label{sec:rw-fusion}
Encoder--decoder models from FCN~\cite{long2015fcn} and U-Net~\cite{unet} to PSPNet~\cite{pspnet}, DeepLabv3+~\cite{deeplabv3plus}, HRNet~\cite{hrnet} and UPerNet~\cite{xie2018uper} established multi-level aggregation as a standard recipe for dense prediction.  The same idea has been specialized for remote sensing through CNN and Transformer decoders such as ABCNet~\cite{abcnet}, BANet~\cite{banet}, MANet~\cite{manet}, A$^2$-FPN~\cite{a2fpn}, UNetFormer~\cite{unetformer}, the FT-UNetFormer implementation in GeoSeg~\cite{ftunetformer} and DCSwin~\cite{dcswin}.  These methods differ mainly in how encoder stages are combined: some use fixed top-down laterals, others add global pooling or channel attention~\cite{senet,cbam,fpn}, and a smaller subset predicts spatially varying weights.

Existing hierarchical decoders therefore already demonstrate the value of multi-level aggregation, while different designs vary in how strongly they adapt fusion to local content.  HG-SAF focuses specifically on dense spatial routing conditioned by a lightweight local heterogeneity statistic.  It is evaluated against mean fusion, globally pooled gates, and content-adaptive pixel-wise alternatives under the same backbone, so that the contribution of the heterogeneity cue can be isolated from the mere use of a pixel-wise gate.

\subsection{Spatial-Frequency Feature Refinement}
\label{sec:rw-freq}
Generic spectral mixers including GFNet~\cite{gfnet}, FNet~\cite{fnet} and FFC~\cite{ffc}, together with Fourier operator learning~\cite{fno}, show that Fourier-domain modulation can complement spatial convolution.  Recent remote sensing models go further.  SFFNet~\cite{sffnet} fuses wavelet and spatial features, FDNet~\cite{fdnet} decouples frequency bands, FGNet~\cite{fgnet} inserts FFT-guided filters into a Swin backbone, AFENet~\cite{afenet} modulates bands according to image content, FreDNet~\cite{frednet} performs frequency-guided denoising, and WgANet~\cite{wganet} uses wavelet-guided attention.  These works establish frequency modeling as a competitive tool for VHR segmentation rather than as a new primitive.

The remaining design question is how a spectral branch should be attached to an already strong fused representation.  \hafr{} does not use frequency modeling to replace the fused spatial features.  FRA restricts frequency modeling to a bounded residual correction inside a channel bottleneck and is directly compared with standard spatial residuals and spectral alternatives at the same insertion point.

\subsection{Structure-Aware Semantic Decoding}
\label{sec:rw-struct}
Boundary supervision~\cite{gscnn,rcf,boundaryloss}, objectness or imbalance losses~\cite{focal,diceloss,lovasz}, and pixel-level contrast or prototype learning~\cite{wang2022contrastive,protoseg} are established tools for sharpening contours and separating similar classes.  Individual boundary, objectness, and prototype mechanisms are not claimed as novel primitives.  CATP uses them as a compact final decoder that follows hierarchical adaptive refinement, and matched controls test whether the combination improves on conventional auxiliary losses.

Recent Transformer and Mamba backbones~\cite{swin,segformer,mask2former,rs3mamba,pyramidmamba,unetmamba} strengthen long-range context, and a parallel line keeps a pretrained model fixed and adapts it with lightweight modules, as in the adapter and LoRA fine-tuning of SAM for multimodal remote sensing segmentation~\cite{mfnet}.  Neither line by itself specifies how the hierarchical features of such an encoder should be organized and refined for VHR scenes.  \hafr{} keeps Swin-B as the representation backbone and places the emphasis on progressive decoding.

\section{Method}
\label{sec:method}

\subsection{Overview}
\label{sec:overview}
\hafr{} is designed around a single principle: progressively refine pretrained hierarchical representations rather than replace them with an increasingly complex task-specific decoder.  Given an input image $\mathbf{I}$, a pretrained Swin-B encoder~\cite{swin} produces four stage features $\{F_s\}_{s=1}^{4}$.  Each stage first passes a fixed preparation block $T_s$, which prepares the stage feature and is kept identical across the module studies of \secref{analysis} (\secref{prep}).  HG-SAF then projects the stages to a common space and adaptively aggregates them into $F_\mathrm{fuse}$.  FRA computes a bounded residual correction to obtain $F_\mathrm{ref}$.  CATP finally maps $F_\mathrm{ref}$ to semantic logits, while auxiliary structural signals are used only during training.

\begin{figure*}[t]
\centering
\includegraphics[width=\linewidth]{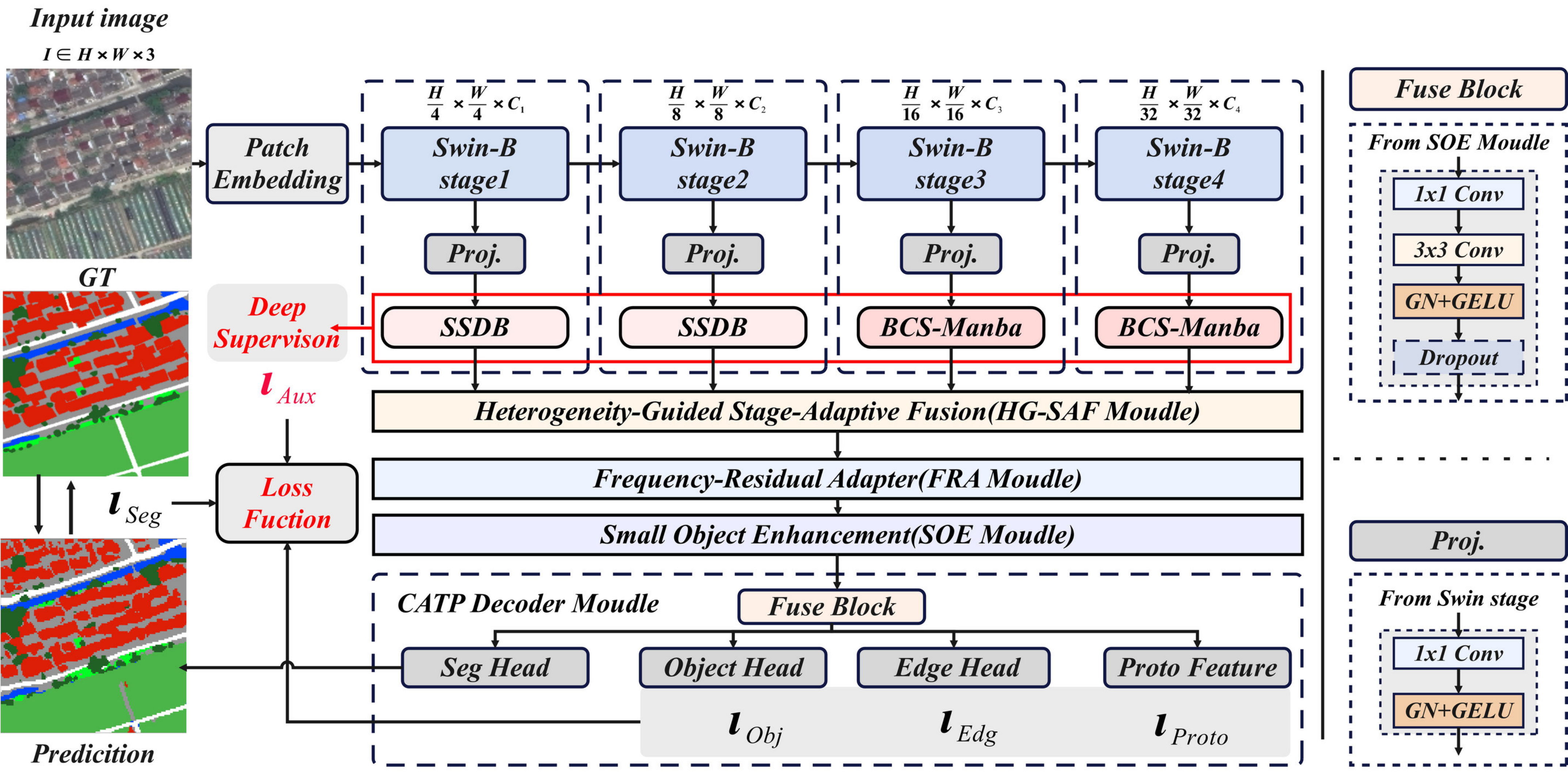}
\caption{Overall architecture of \hafr{}.  A pretrained Swin-B encoder produces four hierarchical stages, which pass fixed preparation blocks (SSDB at the two high-resolution stages, BCS-Mamba at the two low-resolution stages).  HG-SAF then performs pixel-wise adaptive fusion, FRA applies a bounded residual frequency correction, SOE acts on the refined feature, and CATP produces the semantic prediction.  Auxiliary heads are used only in training.}
\label{fig:arch}
\end{figure*}

The encoder is Swin-B pretrained on ImageNet-22k and fine-tuned on ImageNet-1k~\cite{imagenet}, whose four stages have widths $(128,256,512,1024)$; a $1{\times}1$ projection maps them to $(96,192,384,768)$ before the preparation blocks.  Each stage is then aligned to width $C_e{=}384$ and spatial size $(H_1,W_1)=(H/4,W/4)$ before fusion.  As illustrated in \figref{arch}, the inference path is
\begin{equation}
P_\mathrm{sem}=\mathrm{CATP}\Bigl(\mathrm{SOE}\bigl(\mathrm{FRA}\bigl(\mathrm{HG-SAF}(\{T_s(F_s)\}_{s=1}^{4})\bigr)\bigr)\Bigr).
\label{eq:forward}
\end{equation}
The decomposition in \equref{forward} separates three functions that are commonly entangled in a decoder: selecting hierarchical information, refining the aggregated representation, and imposing the tri-prior decoding constraints.  The remainder of this section first records the fixed preparation blocks, then describes the three refinement stages in that order, and closes with their initialization and the training objective.

\subsection{Stage-Specialized Preparation}
\label{sec:prep}
\begin{figure*}[t]
\centering
\includegraphics[width=\linewidth]{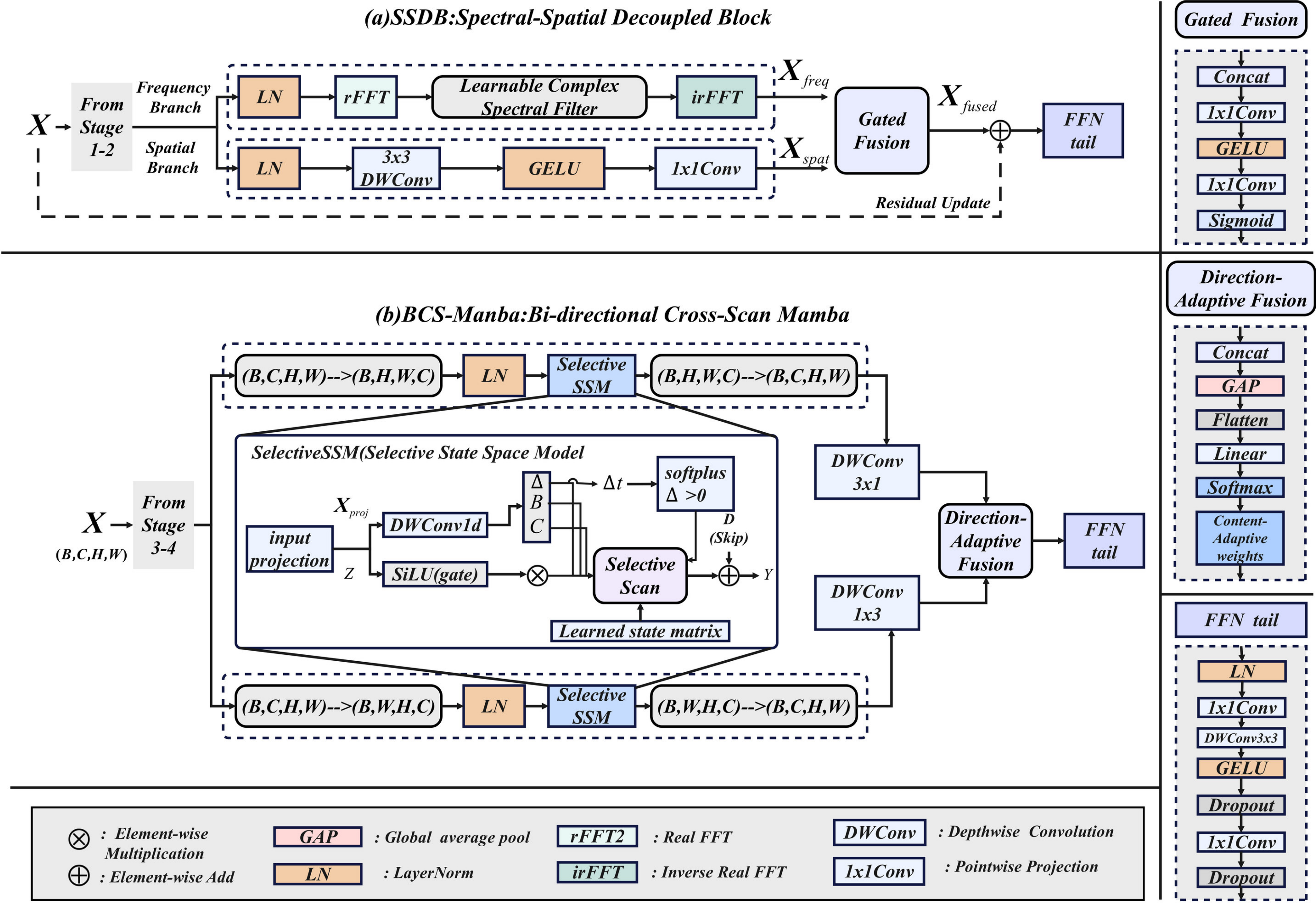}
\caption{Fixed stage preparation blocks, shown with the shared sub-blocks on the right.  (a)~SSDB decouples a normalized stage feature into a learnable complex spectral branch and a depthwise spatial branch, and merges them by a gated residual update.  (b)~BCS-Mamba scans rows and columns as independent selective state-space sequences, restores cross-axis context through two orthogonal depthwise bridges, and merges the two directions by content-adaptive weights.  Both blocks are held identical across the module studies of \secref{analysis}; their measured effect is reported in \secref{trunk}.}
\label{fig:prep}
\end{figure*}

Three blocks are inherited from our implementation and are not claimed as contributions: SSDB and BCS-Mamba prepare the encoder stages, and SOE acts on the refined feature just before decoding.  They are held identical in every module-level comparison of \secref{analysis}, so they cannot account for a difference between the rows of those studies, and \secref{trunk} measures what they contribute to the margin of \tabref{controlled}.  The two stage-preparation blocks are shown in \figref{prep}.

The two high-resolution stages use a spectral--spatial decoupled block (SSDB, \figref{prep}(a)).  A normalized stage feature is processed in parallel by a learnable per-channel complex filter applied to its orthonormal two-dimensional real FFT and by a depthwise $3{\times}3$ convolution followed by a pointwise convolution; a gate conditioned on both branches mixes them as a residual update, and a convolutional feed-forward tail closes the block.  The two low-resolution stages use a bi-directional cross-scan Mamba block (BCS-Mamba, \figref{prep}(b)): rows and columns are scanned as independent sequences, which shortens each scan from $H_1W_1$ to $\max(H_1,W_1)$ steps, two orthogonal depthwise bridges ($3{\times}1$ and $1{\times}3$) re-inject the cross-axis context that axial scanning removes, and a direction-adaptive weighting merges the horizontal and vertical outputs.  The third fixed block, small-object enhancement (SOE), combines parallel dilated depthwise convolutions at rates $2$, $4$ and $8$ with channel attention and a residual scale initialized to $0.1$; it is not a stage-preparation block but acts on $F_\mathrm{ref}$ immediately before decoding, identically in every reported model.  \secref{trunk} measures each of the three on the reference decoder; all three effects are below $+0.2$\,pp, so the gains analyzed in this paper are not produced by these fixed blocks.

\subsection{Heterogeneity-Guided Stage-Adaptive Fusion}
\label{sec:haf}
Hierarchical encoders expose features with progressively lower spatial resolution and stronger semantic abstraction.  In VHR imagery, the relative usefulness of these stages can vary considerably within one tile.  HG-SAF therefore performs dense rather than globally shared stage weighting.

\begin{figure}[t]
\centering
\includegraphics[width=\linewidth]{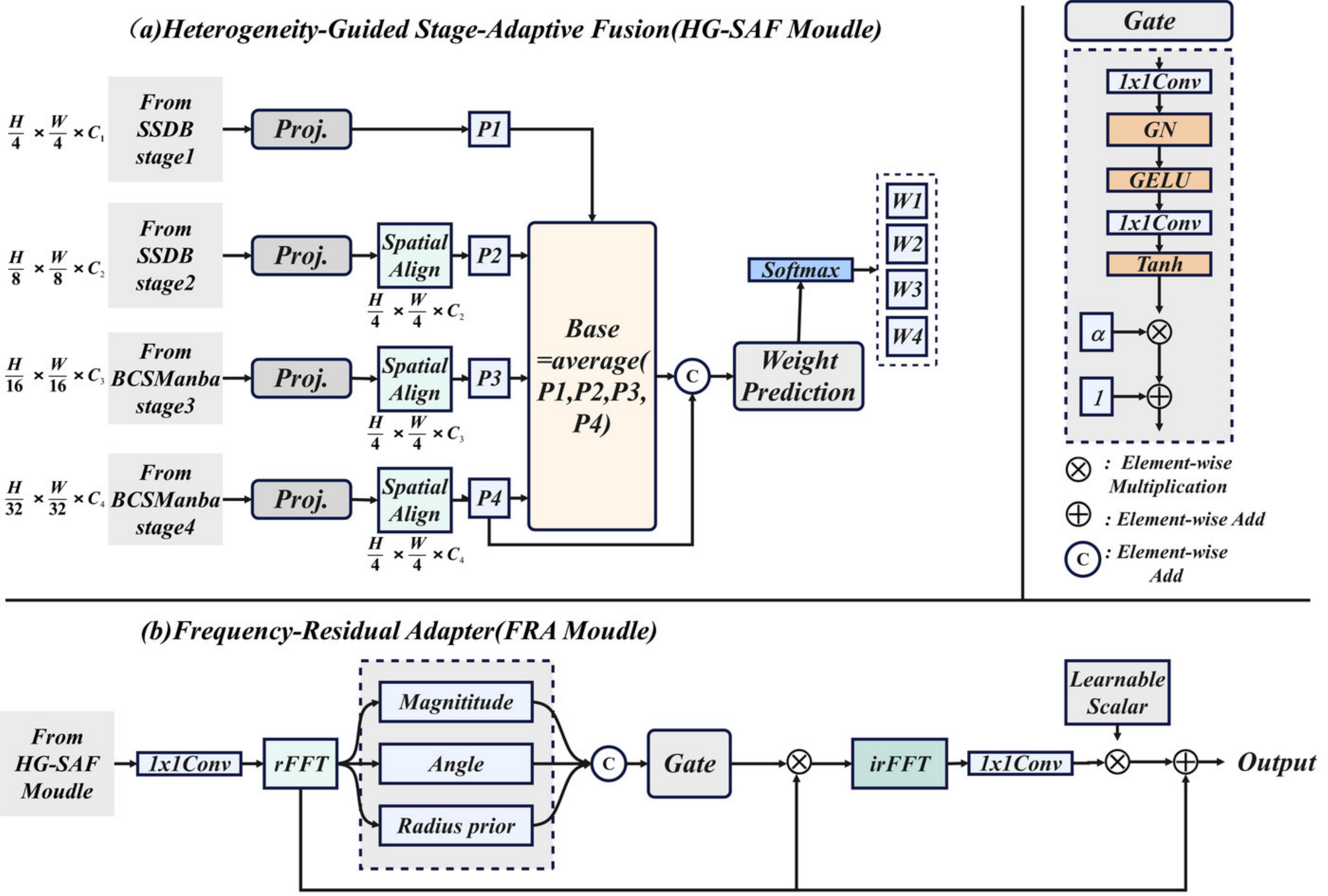}
\caption{The two refinement blocks after the encoder pyramid.  \textbf{(a)}~HG-SAF predicts per-pixel stage weights from $B$ and $\Hetero$ with a zero-initialized gate.  \textbf{(b)}~FRA adds a channel-bottlenecked residual rFFT branch with a tanh-bounded gate and $\gamma{=}0$.}
\label{fig:fig4-hgsaf-fra}
\end{figure}

Each stage is projected to width $C_e$ and upsampled to $(H_1,W_1)$, giving $\tilde F_s$.  Local feature heterogeneity is a lightweight statistic of variation in the deepest projected representation,
\begin{equation}
\Hetero(\bm{x}) = \frac{1}{C_e}\sum_{c=1}^{C_e}\sqrt{\mathrm{Var}_{k\times k}\!\bigl(\tilde F_4^{(c)}\bigr)(\bm{x})+\varepsilon},
\label{eq:hetero}
\end{equation}
with a $k{\times}k$ box filter ($k{=}5$) and $\varepsilon{=}10^{-5}$.  A convolutional gate $g_\theta$, a $3{\times}3$ convolution to $C_e/4$ channels followed by normalization, a GELU and a $1{\times}1$ convolution to the four stage logits, predicts pixel-wise stage weights $\bm{w}=(w_1,\dots,w_4)^\top$, normalized over the stage axis, from the mean stage context $B=\frac{1}{4}\sum_s\tilde F_s$ and $\Hetero$ (\figref{fig4-hgsaf-fra}(a)),
\begin{align}
\bm{w}(\bm{x}) &= \mathrm{softmax}_{s}\!\bigl(g_\theta\!\bigl(\mathrm{cat}[B,\Hetero]\bigr)(\bm{x})\bigr) \in \Delta^{3}\!\subset\!\mathbb{R}^{4}, \label{eq:wsoftmax}\\
F_\mathrm{fuse}(\bm{x}) &= \sum_{s=1}^{4} w_s(\bm{x})\,\tilde F_s(\bm{x}). \label{eq:fuse}
\end{align}
The final projection of $g_\theta$ is zero-initialized, so \equref{wsoftmax} returns uniform stage weights at the first forward pass and \equref{fuse} reduces to the mean of the projected stages.  HG-SAF therefore starts from uniform mean fusion of the projected stages, a non-adaptive reference, and learns spatially varying routing during fine-tuning; the stage projections themselves are trained from scratch.  Matched controls keep the gate architecture and capacity of $g_\theta$ and drop only $\Hetero$ from its input in \equref{wsoftmax}, so that pixel-wise routing and the heterogeneity statistic are not confounded.

\subsection{Frequency-Residual Adapter}
\label{sec:bfr}
After hierarchical fusion, the representation already contains substantial spatial and semantic information.  We therefore avoid replacing it with an unrestricted spectral transformation.  Instead, FRA uses frequency modeling only to construct a bounded residual correction (\figref{fig4-hgsaf-fra}(b)).  With a channel bottleneck of width $r=\max(C_e/4,16)$,
\begin{align}
\widehat Z &= \mathrm{rFFT}\!\bigl(\mathrm{Conv}_{1\!\times\!1}^{C_e\to r}(F_\mathrm{fuse})\bigr), \label{eq:fra-fft}\\
\widetilde Z &= \widehat Z \odot g,\qquad g = 1 + \alpha\tanh\!\bigl(g_\phi(G)\bigr), \label{eq:fra-gate}\\
F_\mathrm{ref} &= F_\mathrm{fuse} + \gamma\,\mathrm{Conv}_{1\!\times\!1}^{r\to C_e}\!\bigl(\mathrm{irFFT}(\widetilde Z;H_1,W_1)\bigr),
\label{eq:fra-out}
\end{align}
Both transforms are orthonormal and act on the half spectrum, so the output of \equref{fra-out} is real by construction.  Here $G$ concatenates the log-amplitude, the sine and cosine of the phase, and a radial prior over the half-spectrum grid, $\alpha{=}0.5$, and $\gamma$ is initialized to $0$.  Three constraints follow from \equrange{fra-fft}{fra-out}.  The bottleneck of width $r$ in \equref{fra-fft} caps both the cost of the spectral branch and the channel rank of its residual mapping.  The multiplier $g$ of \equref{fra-gate} obeys $|g-1|<\alpha$, so the relative rescaling of any spectral coefficient stays inside $(1{-}\alpha,1{+}\alpha)$ and no coefficient can be arbitrarily amplified; the overall correction is additionally scaled by $\gamma$.  The residual scale of \equref{fra-out} makes the first forward pass an exact identity: the branch starts from zero contribution and is learned from data during fine-tuning.  At inference, FRA remains a single residual correction on $F_\mathrm{fuse}$; it does not introduce an auxiliary prediction head.

\subsection{Confusion-Aware Tri-Prior Decoder}
\label{sec:sad}
Even after hierarchical fusion and feature refinement, a single semantic head provides only region-level supervision.  CATP therefore uses lightweight auxiliary signals to regularize boundaries, foreground structure, and inter-class relations during training.

\begin{figure}[t]
\centering
\includegraphics[width=\linewidth]{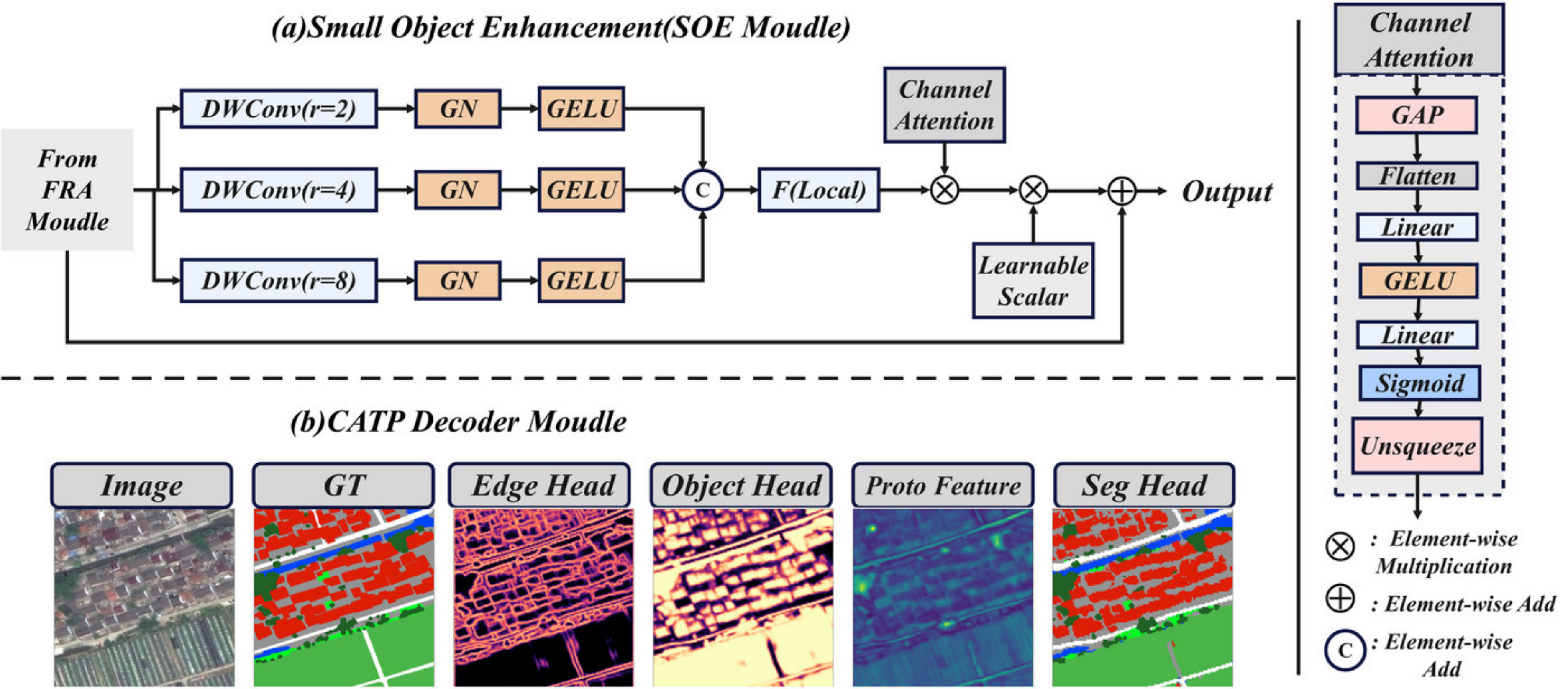}
\caption{Confusion-aware tri-prior decoder.  \textbf{(a)}~The small-object enhancement block (SOE), a fixed block on the fused path rather than a claimed contribution.  \textbf{(b)}~Semantic, boundary, objectness and prototype-feature maps on a Vaihingen tile.  The objectness head predicts a coarse foreground prior, which regularizes $\bar F$ without selecting objects by size.}
\label{fig:fig5-soe-decoder}
\end{figure}

A two-layer fuse block produces $\bar F$.  A boundary branch of two $3{\times}3$ convolutions at width $C_e/4$ followed by a $1{\times}1$ projection predicts single-channel boundary logits $P_\mathrm{edge}$, which are fed back as
\begin{equation}
\bar F' \;=\; \bar F + \beta_e\,\mathrm{enhance}\!\bigl(\mathrm{maxpool}_{3\times3}\!\bigl(\sigma(P_\mathrm{edge})\bigr)\odot \bar F\bigr),
\label{eq:edge-att}
\end{equation}
where $\sigma$ is the logistic sigmoid, broadcast over channels, $\mathrm{enhance}$ is a $3{\times}3$ convolution with normalization and GELU, and $\beta_e$ is initialized to $0.05$.  Three $1{\times}1$ heads applied to $\bar F'$ output semantic logits $P_\mathrm{sem}$, objectness logits $P_\mathrm{obj}$ and a prototype embedding $Q\in\mathbb{R}^{64\times H_1\times W_1}$ (\figref{fig5-soe-decoder}).  The objectness target $T_\mathrm{obj}$ marks the valid pixels outside the two background-like classes of each legend (\textit{Impervious surface} and \textit{Clutter} on ISPRS, \textit{Background} and \textit{Barren} on LoveDA, \textit{Developed} and \textit{Bareland} on OpenEarthMap), so the head learns a coarse foreground prior; small-component IoU is reserved for evaluation.  The edge target is the $4$-neighbour class-difference mask.  Both $\mathcal{L}_\mathrm{edge}$ and $\mathcal{L}_\mathrm{obj}$ are binary CE in which positive pixels carry the weight $\mathrm{clip}(N_-/N_+,1,20)$, where $N_+$ and $N_-$ count positive and negative pixels of the mini-batch.

Some class pairs exhibit substantially larger cross-class errors than others.  We therefore define a small relation set $\mathcal{C}$ from a pilot model using only an internal split of the training partition; the official evaluation split is never used for pair selection.  Per-class prototype $\bm{p}_c$ is the mini-batch mean of $Q$ over the valid pixels of class $c$, re-normalized to unit $\ell_2$ norm so that $\langle\bm{p}_a,\bm{p}_b\rangle$ is a cosine similarity; classes absent from a batch are skipped, and
\begin{equation}
\mathcal{L}_\mathrm{proto} = \frac{1}{|\mathcal{C}|}\!\sum_{(a,b)\in\mathcal{C}}\!\max\!\bigl(0,\;m + \langle\bm{p}_a,\bm{p}_b\rangle\bigr),
\label{eq:proto}
\end{equation}
with $m{=}0.2$, so a pair stops contributing once its prototypes reach $\langle\bm{p}_a,\bm{p}_b\rangle\le-m$.  On the six-class ISPRS label set, $\mathcal{C}$ contains four pairs.  The pair lists for all four datasets are given in the supplement.  At inference, only $P_\mathrm{sem}$ is used.

\subsection{Conservative Initialization of Refinement Branches}
\label{sec:init}
\hafr{} uses conservative initialization to avoid introducing large task-specific perturbations at the beginning of fine-tuning~\cite{rebuffi2017,houlsby2019,layerscale}.  The three components do not share an identical mathematical behavior.  As summarized in \tabref{init}, HG-SAF starts from baseline-equivalent uniform stage fusion, FRA is an exact residual identity when $\gamma{=}0$ in \equref{fra-out}, and the structural feedback path in CATP starts from the small residual coefficient $\beta_e$ of \equref{edge-att}.  We distinguish these cases rather than treating all components as exact identities.

\begin{table}[!b]
\centering
\caption{Conservative Initialization of the Three Refinement Branches}
\label{tab:init}
\renewcommand{\arraystretch}{1.18}
\setlength{\tabcolsep}{5.0pt}
\footnotesize
\arrayrulecolor{rsrule}
\begin{tabularx}{\linewidth}{l l L l}
\tabtoprule
\rowcolor{rshead}
\textbf{Branch} & \textbf{Init.} & \textbf{First forward} & \textbf{Role} \\
\tabheadrule
HG-SAF & last gate $=0$ & uniform $1/4$ fusion & adaptive fusion \\
FRA & $\gamma=0$ & exact identity & residual refine. \\
CATP fb. & $\beta_e{=}0.05$ & near-ref.\ residual & struct.\ feedback \\
\tabbotrule
\end{tabularx}
\arrayrulecolor{black}
\tabnote{The three branches do not share one mathematical identity: HG-SAF starts from mean fusion, FRA is an exact residual identity, and the CATP feedback path is a small residual.}
\end{table}

These three reference behaviors are recorded so that later ablations can isolate initialization from architecture.  The next subsection defines the training objective on the semantic head and the auxiliary structural signals.

\subsection{Training Objective}
\label{sec:loss}
Let $\Omega=\{i:y_i\neq255\}$ be the valid pixels, so that ignore index $255$ keeps unlabeled and eroded-boundary pixels out of $\mathcal{L}_\mathrm{CE}$ and $\mathcal{L}_\mathrm{FocalDice}$.  Let $\mathrm{DSC}_c=(2\sum_{i\in\Omega} p_{i,c}y_{i,c}+\varepsilon_\mathrm{d})/(\sum_{i\in\Omega} p_{i,c}+\sum_{i\in\Omega} y_{i,c}+\varepsilon_\mathrm{d})$ with $\varepsilon_\mathrm{d}{=}1$, one-hot targets $y_{i,c}$ and $p=\mathrm{softmax}_c(P_\mathrm{sem})$.  The semantic loss is $\mathcal{L}_\mathrm{sem}=\mathcal{L}_\mathrm{CE}+\lambda_d\mathcal{L}_\mathrm{FocalDice}$, where
\begin{equation}
\mathcal{L}_\mathrm{FocalDice}=\frac{1}{N_\mathrm{cls}}\sum_{c=1}^{N_\mathrm{cls}}\bigl(1-\mathrm{DSC}_c\bigr)^{\kappa},
\label{eq:focaldice}
\end{equation}
$N_\mathrm{cls}$ is the number of scored classes, $\kappa{=}2.5$, and $\mathcal{L}_\mathrm{CE}$ uses class-balanced weights (\textit{Car}~$3{\times}$, \textit{Clutter}~$5{\times}$ on ISPRS).  The full objective is
\begin{equation}
\mathcal{L}=\mathcal{L}_\mathrm{sem}+\lambda_e\mathcal{L}_\mathrm{edge}+\lambda_o\mathcal{L}_\mathrm{obj}+\lambda_p\mathcal{L}_\mathrm{proto}+\mathcal{L}_\mathrm{aux},
\label{eq:loss}
\end{equation}
with $\lambda_d{=}1.0$, $\lambda_e{=}0.4$, $\lambda_o{=}0.2$ and $\lambda_p{=}0.1$.  The auxiliary term $\mathcal{L}_\mathrm{aux}$ attaches CE and FocalDice heads to the prepared stages $\{T_s(F_s)\}_{s=1}^{4}$ with weights $(0.4,0.3,0.2,0.1)$.  Every term of \equref{loss} is evaluated at the label resolution: the auxiliary and structural logits are bilinearly upsampled to the input size, so the discrete label map is never interpolated and keeps its ignore index.  Only the class map of $\mathcal{L}_\mathrm{proto}$ is resampled, to $(H_1,W_1)$ and by nearest neighbour.  All auxiliary heads are disabled at inference.  The next section specifies the datasets, matched protocol, and metrics used to evaluate this design.

\section{Experimental Setup}
\label{sec:setup}

\subsection{Datasets and Exact Splits}
\label{sec:data}
\textbf{ISPRS Vaihingen}~\cite{vaihingen_isprs} contains $33$ NIR-RG tiles at $9$\,cm GSD.  The $16$ official training tiles are 1, 3, 5, 7, 11, 13, 15, 17, 21, 23, 26, 28, 30, 32, 34 and 37; the remaining $17$ tiles form the test set, stored as $113$ non-overlapping $1024{\times}1024$ patches, which are the evaluation and bootstrap unit.  Five foreground classes are scored; \textit{Clutter} is excluded; official $3$\,px boundary erosion and ignore index $255$ are used.

\textbf{ISPRS Potsdam}~\cite{potsdam_isprs} contains $38$ RGB tiles at $5$\,cm GSD and the same six-class legend.  Train tiles: $2\_10$--$2\_12$, $3\_10$--$3\_12$, $4\_10$--$4\_12$, $5\_10$--$5\_12$, $6\_7$--$6\_12$, $7\_7$--$7\_12$.  Test tiles: $2\_13$, $2\_14$, $3\_13$, $3\_14$, $4\_13$--$4\_15$, $5\_13$--$5\_15$, $6\_13$--$6\_15$, $7\_13$.  The same five-class eroded protocol is used.

\textbf{LoveDA}~\cite{loveda} contains $5987$ RGB images at $30$\,cm GSD and seven classes.  We report the official validation split.  Several published Transformer and Mamba numbers were originally submitted to the online test server; they appear only as contextual literature values.

\textbf{OpenEarthMap}~\cite{openearthmap} contains $5000$ images at $25$--$50$\,cm GSD.  mIoU averages the eight official land-cover classes; unlabeled pixels are ignored.  There is no separate Background class.

Every image enters the network at $512{\times}512$: the ISPRS tiles are held as the non-overlapping $1024{\times}1024$ patches listed above and each patch is resampled to that size, as are the LoveDA and OpenEarthMap images, which the two benchmarks already distribute as single tiles.  Labels are resampled by nearest neighbour so that class indices stay valid, and the ignore index is preserved.  Normalization uses ImageNet mean and standard deviation.  No CRF, multi-crop merging, or other post-processing is applied.

\subsection{Controlled Baselines and Published Comparisons}
\label{sec:baselines}
Conclusions are drawn from a controlled Swin-B comparison: UPerNet~\cite{xie2018uper}, FPN~\cite{fpn}, DeepLabv3+~\cite{deeplabv3plus}, SegFormer MLP~\cite{segformer}, a UNetFormer-style decoder~\cite{unetformer}, SE-gated fusion~\cite{senet} and an FFT mixer~\cite{gfnet}, all retrained with the single recipe of \secref{impl}.  Published CNN, Transformer and Mamba numbers are provided for contextual comparison and are not treated as strictly controlled comparisons, because training, backbone, metric, and inference configurations differ across sources.  FT-UNetFormer is cited as the GeoSeg implementation~\cite{ftunetformer}.  A provenance table is given in the supplement.

\subsection{Implementation Details}
\label{sec:impl}
Training uses PyTorch~2.1~\cite{pytorch} and one RTX~4090.  Every controlled model, including \hafr{}, shares one recipe: $512{\times}512$ inputs, at most $80$ epochs with batch size $8$, AdamW~\cite{adamw} with head learning rate $1.2{\times}10^{-4}$ and backbone learning rate $4{\times}10^{-5}$, a cosine schedule with a $5$-epoch warmup~\cite{warmup}, weight decay $10^{-4}$, gradient clipping $1.0$ and BF16 mixed precision.  Augmentation is horizontal or vertical flip and discrete $90^\circ$ rotation, each with probability $0.5$.  The auxiliary deep supervision of \equref{loss} belongs to this shared recipe: every controlled model carries CE and FocalDice heads on the corresponding encoder stages with the weights $(0.4,0.3,0.2,0.1)$, so \tabref{controlled} compares decoding paths and not the amount of training signal.  Seeds are $42$, $43$ and $44$, and the reported checkpoint is the best-mIoU epoch under one monitoring rule that is identical for every controlled model.

Controlled inference is single-scale without test-time augmentation (TTA).  For the two ISPRS sets we additionally report flip and $\{90^\circ,180^\circ,270^\circ\}$ TTA so that TTA gain can be separated from method gain.  LoveDA is the only dataset with a stronger photometric recipe: color jitter $0.4$, Gaussian blur $0.3$, EMA with decay $0.999$~\cite{tarvainen2017ema} and early stopping with patience $12$; ISPRS and OpenEarthMap use none of these.  This difference is recorded rather than hidden, and no controlled baseline on LoveDA is trained under a weaker recipe than \hafr{}.

\subsection{Evaluation Metrics}
\label{sec:metrics}
We report mIoU and per-class IoU.  HG-SAF is further examined through heterogeneity- and size-stratified IoU and stage-weight statistics; the strata are computed once from the reference model and reused for every variant, so all rows are scored on identical pixel sets.  FRA is examined through Boundary IoU~\cite{boundaryiou}, boundary F-score~\cite{bfscore}, thin-structure IoU (local width below $8$\,px) and small-component IoU (area quartile Q1).  CATP is examined through pairwise confusion mass $E_{a,b}=M_{ab}+M_{ba}$ on the pre-declared relation set, where $M$ is the confusion matrix normalized by the number of valid pixels.  Uncertainty is a paired bootstrap ($1\,000$ resamples) of the mIoU difference versus UPerNet over the non-overlapping evaluation patches, so no pixel enters a replicate twice.  Efficiency uses input $512{\times}512$, batch size $1$, BF16, $50$ warm-up iterations and $500$ timed iterations with \texttt{torch.cuda.synchronize()}; the timed loop excludes host-device copy and post-processing.

\section{Experimental Results}
\label{sec:results}

\subsection{Controlled Comparison}
\label{sec:controlled}
\tabref{controlled} reports the controlled comparison under a matched training and inference protocol, and the conclusions of this paper are drawn from it.  It holds the encoder, the schedule and the inference setting fixed, so that a difference between two rows is a difference in the decoding path rather than in training or inference.  \hafr{} carries the fixed blocks of \secref{prep} in addition to its three refinement stages, and \secref{trunk} measures their separate share of the margin.

\begin{table}[!t]
\centering
\caption{Controlled Comparison of Decoders Under a Matched Protocol}
\label{tab:controlled}
\renewcommand{\arraystretch}{1.16}
\setlength{\tabcolsep}{4.6pt}
\footnotesize
\arrayrulecolor{rsrule}
\begin{tabularx}{\linewidth}{L ccccc}
\tabtoprule
\rowcolor{rshead}
\textbf{Method (Swin-B)} & \textbf{Vaih.} & \textbf{Pots.} & \textbf{LoveDA} & \textbf{OEM} & \textbf{Avg.} \\
\tabheadrule
UPerNet~\cite{xie2018uper} & 83.57 & 86.91 & 53.62 & 65.86 & 72.49 \\
FPN~\cite{fpn} & 83.21 & 86.54 & 53.18 & 65.41 & 72.09 \\
DeepLabv3+~\cite{deeplabv3plus} & 83.38 & 86.72 & 53.35 & 65.58 & 72.26 \\
SegFormer MLP~\cite{segformer} & 83.44 & 86.80 & 53.48 & 65.71 & 72.36 \\
UNetFormer dec.~\cite{unetformer} & \secv{83.79} & \secv{87.14} & \secv{54.02} & \secv{66.18} & \secv{72.78} \\
SE-gated fusion~\cite{senet} & 83.68 & 87.05 & 53.88 & 66.02 & 72.66 \\
FFT spectral mixer~\cite{gfnet} & 83.62 & 86.98 & 53.71 & 65.94 & 72.56 \\
\rowcolor{rsmark}
\oursrow{\hafr{} (no TTA)} & \bestv{84.12} & \bestv{87.86} & \bestv{55.17} & \bestv{67.70} & \bestv{73.71} \\
\tabheadrule
\multicolumn{6}{l}{\textit{TTA (flip$+$rotation), reported only for the two ISPRS sets}}\\
UPerNet$^\dagger$ & \secv{83.89} & \secv{87.24} & -- & -- & -- \\
\rowcolor{rsmark}
\oursrow{\hafr{}$^\dagger$} & \bestv{84.48} & \bestv{88.20} & -- & -- & -- \\
\tabbotrule
\end{tabularx}
\arrayrulecolor{black}
\tabnote{All models share the split, crop, optimizer, schedule, augmentation, and checkpoint rule of \secref{impl}, and use single-scale inference without TTA unless marked $^\dagger$.  Entries are the mean mIoU (\%) over seeds $42$, $43$, and $44$.  The paired bootstrap of \secref{metrics} gives $95\%$ confidence intervals of $[0.31,0.79]$, $[0.66,1.22]$, $[1.18,1.91]$ and $[1.47,2.19]$\,pp for the no-TTA margin of \hafr{} over UPerNet on the four datasets, so every margin excludes zero.  \bestsec{}  The no-TTA and TTA blocks are ranked separately.}
\end{table}

Under the no-TTA protocol, \hafr{} improves Swin-B$+$UPerNet by $+0.55$\,pp on Vaihingen ($95\%$ CI $[0.31,0.79]$), $+0.95$\,pp on Potsdam ($[0.66,1.22]$), $+1.55$\,pp on LoveDA ($[1.18,1.91]$), and $+1.84$\,pp on OpenEarthMap ($[1.47,2.19]$); all four intervals exclude zero, and the ordering is preserved across the three seeds.  The gains are larger on the more heterogeneous LoveDA and OpenEarthMap sets than on the saturated ISPRS urban tiles.  Among the controlled decoders, the UNetFormer-style decoder is the strongest alternative, and \hafr{} exceeds it by $+0.33$ to $+1.52$\,pp across the four sets.  TTA adds a further $+0.36$ / $+0.34$\,pp on the two ISPRS sets.  The improvement holds on all four datasets under a single protocol, so it follows from the decoding design rather than from a dataset-specific trick.

\subsection{Contextual Comparison with Published Results}
\label{sec:sota}
For broader context, we additionally report representative published results from CNN-, Transformer-, and Mamba-based remote sensing models.  Because these values are obtained under different backbones, training recipes, and inference settings, they are treated as contextual references rather than strict controlled comparisons.  Under this reading, \hafr{} attains the best or the second-best reported accuracy on each of the four benchmarks.

\begin{table*}[!t]
    \centering
    \caption{Contextual Comparison With Published Results on ISPRS Vaihingen and Potsdam}
    \label{tab:isprs}
    \renewcommand{\arraystretch}{1.16}
    \setlength{\tabcolsep}{6.4pt}
    \footnotesize
    \arrayrulecolor{rsrule}
    \begin{tabularx}{\linewidth}{L cccccc cccccc}
    \tabtoprule
    \rowcolor{rshead}
    & \multicolumn{6}{c}{\textbf{ISPRS Vaihingen}} & \multicolumn{6}{c}{\textbf{ISPRS Potsdam}} \\
    \cmidrule(lr){2-7}\cmidrule(l){8-13}
    \rowcolor{rshead}
    \textbf{Method} & \textbf{Imp.S.} & \textbf{Build.} & \textbf{L.Veg.} & \textbf{Tree} & \textbf{Car} & \textbf{mIoU}
                    & \textbf{Imp.S.} & \textbf{Build.} & \textbf{L.Veg.} & \textbf{Tree} & \textbf{Car} & \textbf{mIoU} \\
    \tabheadrule
    \multicolumn{13}{l}{\textit{CNN-based}} \\
    UNet~\cite{unet}            & 90.1 & 88.0 & 69.1 & 78.4 & 64.5 & 78.02 & 82.8 & 89.9 & 73.8 & 75.6 & 83.2 & 81.06 \\
    PSPNet~\cite{pspnet}        & 91.0 & 88.7 & 69.5 & 79.1 & 66.8 & 79.02 & 84.2 & 91.0 & 74.8 & 76.5 & 85.2 & 82.34 \\
    DenseASPP~\cite{denseaspp}  & 90.3 & 87.6 & 68.4 & 77.9 & 62.5 & 77.34 & 83.3 & 90.3 & 73.5 & 75.3 & 83.8 & 81.24 \\
    MANet~\cite{manet}          & 91.2 & 89.2 & 70.1 & 79.4 & 68.9 & 79.76 & 84.8 & 91.8 & 75.4 & 77.1 & 86.2 & 83.06 \\
    HRNet~\cite{hrnet}          & 91.6 & 89.6 & 70.5 & 79.9 & 69.9 & 80.30 & 83.8 & 90.7 & 74.2 & 76.1 & 84.5 & 81.86 \\
    ABCNet~\cite{abcnet}        & 90.7 & 88.5 & 69.3 & 78.9 & 67.6 & 79.00 & 84.3 & 91.2 & 75.0 & 76.7 & 85.5 & 82.54 \\
    BANet~\cite{banet}          & 89.8 & 87.3 & 68.1 & 77.8 & 65.0 & 77.60 & 83.5 & 90.5 & 73.9 & 75.7 & 84.2 & 81.56 \\
    A$^2$-FPN~\cite{a2fpn}      & 91.4 & 89.4 & 70.4 & 79.5 & 69.5 & 80.04 & 85.2 & 92.1 & 75.9 & 77.5 & 86.6 & 83.46 \\
    CGNet~\cite{cgnet_rs}       & 90.5 & 88.1 & 68.9 & 78.6 & 67.2 & 78.66 & 84.0 & 90.8 & 74.5 & 76.4 & 85.0 & 82.14 \\
    \multicolumn{13}{l}{\textit{Transformer-based}} \\
    FT-UNetFormer$^\dagger$~\cite{ftunetformer} & 91.9 & 89.9 & 71.1 & 80.3 & 73.3 & 81.30 & 86.2 & 92.9 & 76.8 & 78.3 & 88.0 & 84.44 \\
    DCSwin$^\dagger$~\cite{dcswin}             & 92.7 & 90.5 & 71.7 & 80.8 & 75.1 & 82.16 & 88.0 & 94.2 & 78.6 & 80.8 & 91.9 & 86.70 \\
    Mask2Former$^\dagger$~\cite{mask2former}   & 92.9 & 90.7 & 72.1 & 81.2 & 75.4 & 82.46 & 82.8 & 94.0 & 79.2 & \bestv{82.8} & 73.2 & 82.40 \\
    LOGCAN++~\cite{logcanpp}                   & 92.1 & 90.3 & 72.0 & 81.2 & 77.1 & 82.54 & 87.6 & 93.8 & 77.9 & 80.0 & 89.9 & 85.84 \\
    CG-Swin$^\dagger$~\cite{cgswin}            & 93.1 & 90.8 & 72.4 & 81.3 & 77.6 & 83.04 & 88.7 & \secv{94.9} & 79.4 & 81.3 & \bestv{93.4} & 87.54 \\
    UNetFormer$^\dagger$~\cite{unetformer}     & 93.4 & 91.2 & 72.6 & 81.8 & 78.1 & 83.42 & 86.9 & 93.4 & 77.5 & 79.2 & 88.7 & 85.14 \\
    MMT$^\dagger$~\cite{mmt}                   & 93.4 & 91.4 & 72.9 & 81.8 & 78.7 & 83.64 & 89.6 & 93.8 & 78.9 & 81.2 & 92.7 & 87.24 \\
    \multicolumn{13}{l}{\textit{Mamba-based}} \\
    RS3Mamba$^\dagger$~\cite{rs3mamba}         & 92.5 & 89.9 & 72.1 & 81.0 & 76.2 & 82.34 & 87.3 & 93.6 & 77.8 & 79.5 & 90.5 & 85.74 \\
    UNetMamba$^\dagger$~\cite{unetmamba}       & 92.9 & 90.8 & 72.7 & 81.2 & 76.8 & 82.88 & 88.3 & 94.4 & 78.7 & 80.7 & 92.1 & 86.84 \\
    D2LS$^\dagger$~\cite{d2ls}                 & 93.6 & 91.6 & 73.3 & \secv{82.0} & 79.1 & 83.92 & 89.2 & \secv{94.9} & 79.2 & 81.7 & \secv{93.1} & 87.62 \\
    PyramidMamba$^\dagger$~\cite{pyramidmamba} & \bestv{94.2} & \bestv{92.5} & \secv{74.7} & \bestv{82.3} & \bestv{80.5} & \bestv{84.84} & \secv{89.9} & 94.6 & \secv{79.5} & 81.7 & \bestv{93.4} & \secv{87.82} \\
    \rowcolor{rsmark}
    \oursrow{\hafr{}$^\dagger$ (Ours)} & \secv{93.8} & \secv{92.1} & \bestv{75.0} & \secv{82.0} & \secv{79.5} & \secv{84.48}\std{0.11}
                                       & \bestv{90.7} & \bestv{95.1} & \bestv{80.2} & \secv{82.1} & 92.9 & \bestv{88.20}\std{0.09} \\
    \tabbotrule
    \end{tabularx}
    \arrayrulecolor{black}
    \tabnote{Per-class IoU and mIoU (\%) on the five scored foreground classes (Imp.S.: impervious surface; Build.: building; L.Veg.: low vegetation); \textit{Clutter} is excluded and the official $3$-px boundary erosion is applied.  Methods marked $^\dagger$ use TTA as reported by their sources.  Because backbones, splits, and inference settings differ across papers, these values are literature context rather than a matched benchmark; the matched Swin-B comparison is \tabref{controlled}.  \bestsec{}  Best and second best are ranked separately within each dataset.  The $\pm$ value is the three-seed standard deviation of \hafr{}.}
    \end{table*}

\tabref{isprs} summarizes the two ISPRS urban benchmarks.  With TTA, \hafr{} reaches $84.48\%$ mIoU on Vaihingen and $88.20\%$ on Potsdam.  The per-class pattern is consistent with the later analyses: on Vaihingen \hafr{} leads only on \textit{Low Vegetation}, and on Potsdam its largest margins are on \textit{Impervious surface} ($+0.8$\,pp) and \textit{Low Vegetation} ($+0.7$\,pp).  \hafr{} holds the highest Potsdam mIoU of the table; on Vaihingen only PyramidMamba reports a higher mean, by $0.36$\,pp, with a different backbone and TTA setting.

\begin{table*}[!t]
    \centering
    \caption{Contextual Comparison With Published Results on the LoveDA Validation Split}
    \label{tab:loveda}
    \renewcommand{\arraystretch}{1.16}
    \setlength{\tabcolsep}{12.5pt}
    \footnotesize
    \arrayrulecolor{rsrule}
    \begin{tabularx}{\linewidth}{L ccccccc c}
    \tabtoprule
    \rowcolor{rshead}
    \textbf{Method} & \textbf{Backgr.} & \textbf{Building} & \textbf{Road} & \textbf{Water} & \textbf{Barren} & \textbf{Forest} & \textbf{Agricult.} & \textbf{mIoU} \\
    \tabheadrule
    \multicolumn{9}{l}{\textit{CNN-based}} \\
    UNet~\cite{unet}        & 50.24 & 59.77 & 52.79 & 57.02 & 31.08 & 41.73 & 51.81 & 49.20 \\
    PSPNet~\cite{pspnet}    & \bestv{52.50} & 59.46 & 51.15 & 61.10 & 27.46 & 37.48 & 50.32 & 48.49 \\
    DenseASPP~\cite{denseaspp}& 50.77 & 55.67 & 53.08 & 55.49 & 23.99 & 35.24 & 43.54 & 45.40 \\
    MANet~\cite{manet}       & 49.82 & 58.08 & 52.98 & 54.30 & 25.21 & 37.65 & 51.87 & 47.13 \\
    HRNet~\cite{hrnet}       & 49.90 & 57.33 & 55.94 & 52.34 & 31.64 & 41.94 & 48.50 & 48.23 \\
    A$^2$-FPN~\cite{a2fpn}    & 47.73 & 59.19 & 52.20 & 64.08 & \secv{32.09} & 37.96 & 53.00 & 49.46 \\
    \multicolumn{9}{l}{\textit{Transformer-based}} \\
    LOGCAN++~\cite{logcanpp}        & 47.37 & 58.38 & 56.46 & 80.05 & 18.44 & \secv{47.91} & \bestv{64.80} & 53.35 \\
    AerialFormer~\cite{aerialformer}& 47.80 & \secv{60.70} & \bestv{59.30} & 81.50 & 17.90 & 47.90 & 64.00 & 54.10 \\
    SFA-Net~\cite{sfanet}           & 48.40 & 60.30 & \secv{59.10} & \bestv{81.90} & 24.10 & 46.20 & 64.00 & 54.90 \\
    \multicolumn{9}{l}{\textit{Mamba-based}} \\
    RS3Mamba~\cite{rs3mamba}        & 41.60 & 58.23 & 54.03 & 77.34 & 17.97 & 43.81 & 61.37 & 50.62 \\
    UNetMamba~\cite{unetmamba}      & 47.08 & 59.16 & 56.74 & 81.37 & 18.15 & 46.61 & \secv{64.31} & 53.35 \\
    PyramidMamba~\cite{pyramidmamba}& 51.07 & 56.39 & 52.44 & 66.81 & 28.95 & 35.13 & 45.84 & 48.09 \\
    D2LS~\cite{d2ls}                & 47.60 & \bestv{61.20} & \secv{59.10} & \secv{81.60} & 23.80 & \bestv{48.80} & \bestv{64.80} & \bestv{55.30} \\
    \rowcolor{rsmark}\oursrow{\hafr{} (Ours)}  & \secv{52.32} & 60.16 & 58.80 & 70.38 & \bestv{39.80} & 42.67 & 62.05 & \secv{55.17}\std{0.12} \\
    \tabbotrule
    \end{tabularx}
    \arrayrulecolor{black}
    \tabnote{Per-class IoU and mIoU (\%) on the seven LoveDA classes.  \hafr{} is evaluated on the official validation split with single-scale inference and no TTA.  Several published Transformer and Mamba entries were originally submitted to the online test server and are therefore not strictly comparable; they are shown only as literature context, and the matched Swin-B comparison is \tabref{controlled}.  \bestsec{}  The $\pm$ value is the three-seed standard deviation of \hafr{}; per-class standard deviations do not exceed $0.18$ and are omitted for readability.}
    \end{table*}

\tabref{loveda} reports LoveDA on the official validation split without TTA.  \hafr{} attains $55.17\%$ mIoU.  The largest per-class margin is on \textit{Barren}, the class that the training-only relation set $\mathcal{C}$ of \equref{proto} pairs with \textit{Background} on this dataset.  Only D2LS reports a higher mean, by $0.13$\,pp, and part of the table was produced on the online test server rather than on the validation split, as recorded in the supplement.

\begin{table*}[!t]
    \centering
    \caption{Contextual Comparison With Published Results on OpenEarthMap}
    \label{tab:oem}
    \renewcommand{\arraystretch}{1.16}
    \setlength{\tabcolsep}{10.0pt}
    \footnotesize
    \arrayrulecolor{rsrule}
    \begin{tabularx}{\linewidth}{L ccccccccc}
    \tabtoprule
    \rowcolor{rshead}
    \textbf{Method} & \textbf{Bareland} & \textbf{Range.} & \textbf{Develop.} & \textbf{Road} & \textbf{Tree} & \textbf{Water} & \textbf{Agricult.} & \textbf{Building} & \textbf{mIoU} \\
    \tabheadrule
    \multicolumn{10}{l}{\textit{CNN-based}} \\
    UNet~\cite{unet}          & 33.6 & 49.2 & 49.8 & 56.1 & 64.7 & 71.4 & 70.9 & 72.3 & 58.50 \\
    PSPNet~\cite{pspnet}      & 35.2 & 50.1 & 50.6 & 57.4 & 65.8 & 72.6 & 71.8 & 73.0 & 59.56 \\
    DenseASPP~\cite{denseaspp}& 32.9 & 48.4 & 49.1 & 55.6 & 63.9 & 70.7 & 70.1 & 71.6 & 57.79 \\
    MANet~\cite{manet}        & 36.0 & 50.8 & 51.2 & 58.0 & 66.5 & 73.4 & 72.5 & 73.8 & 60.27 \\
    HRNet~\cite{hrnet}        & 35.6 & 50.5 & 50.9 & 57.7 & 66.1 & 73.0 & 72.1 & 73.4 & 59.91 \\
    ABCNet~\cite{abcnet}      & 34.4 & 49.7 & 50.2 & 56.9 & 65.3 & 72.1 & 71.4 & 72.7 & 59.09 \\
    BANet~\cite{banet}        & 33.0 & 48.6 & 49.4 & 55.8 & 64.2 & 70.9 & 70.4 & 71.9 & 58.03 \\
    A$^2$-FPN~\cite{a2fpn}    & 36.4 & 51.0 & 51.5 & 58.3 & 66.8 & 73.7 & 72.8 & 74.1 & 60.57 \\
    CGNet~\cite{cgnet_rs}     & 34.9 & 49.9 & 50.4 & 57.1 & 65.6 & 72.4 & 71.6 & 72.9 & 59.35 \\
    \multicolumn{10}{l}{\textit{Transformer-based}} \\
    SegFormer$^\dagger$~\cite{segformer}      & 41.0 & 56.4 & 53.2 & 58.7 & 70.9 & 77.7 & 76.7 & 76.2 & 63.85 \\
    FT-UNetFormer$^\dagger$~\cite{ftunetformer}& 41.8 & 55.0 & 52.6 & 57.6 & 69.4 & 77.2 & 75.5 & 75.1 & 63.02 \\
    DCSwin$^\dagger$~\cite{dcswin}            & 42.5 & 55.6 & 53.2 & 58.1 & 69.9 & 77.8 & 76.0 & 75.7 & 63.60 \\
    Mask2Former$^\dagger$~\cite{mask2former}  & 42.6 & 56.0 & 53.9 & 59.8 & 70.0 & 77.5 & 76.3 & 76.4 & 64.06 \\
    UNetFormer$^\dagger$~\cite{unetformer}    & 42.8 & 56.2 & 53.5 & 60.9 & 70.2 & 77.4 & 76.6 & 76.9 & 64.31 \\
    LOGCAN++~\cite{logcanpp}                  & 43.6 & 57.8 & 55.4 & 62.1 & 70.8 & 78.9 & 77.6 & 77.0 & 65.40 \\
    \multicolumn{10}{l}{\textit{Mamba-based}} \\
    RS3Mamba$^\dagger$~\cite{rs3mamba}        & 39.9 & 51.0 & 48.7 & 56.9 & 66.8 & 74.4 & 75.0 & 71.4 & 60.51 \\
    UNetMamba$^\dagger$~\cite{unetmamba}      & 43.2 & 57.1 & 54.8 & 61.4 & 70.5 & 78.3 & 77.1 & 77.2 & 64.95 \\
    PyramidMamba$^\dagger$~\cite{pyramidmamba}& \secv{45.0} & \secv{59.4} & \secv{57.9} & \secv{64.9} & \bestv{72.1} & \bestv{81.3} & \secv{79.4} & \secv{79.6} & \secv{67.45} \\
    \rowcolor{rsmark}\oursrow{\hafr{} (Ours)} & \bestv{45.6} & \bestv{59.8} & \bestv{58.4} & \bestv{65.3} & \secv{71.9} & \secv{80.8} & \bestv{79.7} & \bestv{80.1} & \bestv{67.70}\std{0.10} \\
    \tabbotrule
    \end{tabularx}
    \arrayrulecolor{black}
    \tabnote{Per-class IoU and mIoU (\%) over the eight official land-cover classes; unlabeled pixels are ignored and are not counted as a Background class.  Methods marked $^\dagger$ use TTA as reported by their sources, whereas \hafr{} uses single-scale inference.  These literature numbers are not a matched protocol; see \tabref{controlled} for the controlled comparison.  \bestsec{}  The $\pm$ value is the three-seed standard deviation of \hafr{}; per-class standard deviations do not exceed $0.13$ and are omitted for readability.}
    \end{table*}

\tabref{oem} reports OpenEarthMap under the eight-class official protocol.  The no-TTA model reaches $67.70\%$ mIoU.  Among the pre-declared relation pairs, the largest reduction of confusion mass is on \textit{Rangeland}\,$\leftrightarrow$\,\textit{Agriculture} (\tabref{sad}, \emph{Panel~B}).  Across the three tables \hafr{} therefore holds the best published mean on Potsdam and OpenEarthMap and trails the best entry by $0.36$ and $0.13$\,pp on Vaihingen and LoveDA, and \tabref{controlled} is the matched evidence behind that standing.

\section{Analysis and Ablation}
\label{sec:analysis}

The previous section reports overall accuracy.  This section asks a narrower question: whether each refinement stage behaves in the way it was designed to behave.  We keep the Swin-B encoder fixed and swap only the component under test, and we close with visual examples of the strata that the analyses identify.  Unless a table note states otherwise, ISPRS scores in this section follow the flip-and-rotation protocol of \tabref{isprs}, so the full model is the $84.48\%$ entry of \tabref{controlled}.

\subsection{Stage-Adaptive Fusion}
\label{sec:c1}
\tabref{haf} isolates the contribution of HG-SAF.  Every row of \emph{Panel~A} replaces only the fusion stage, so the ladder decomposes the $+0.20$\,pp that HG-SAF gains over mean fusion: a globally pooled gate recovers $+0.04$, a pixel-wise gate that sees only the mean context $B$ recovers $+0.06$, and the heterogeneity statistic of \equref{hetero} supplies the remaining $+0.14$.  The cue is therefore the larger part of the effect, and it is the statistic rather than the gate that carries it: with the same gate and capacity, the input-gradient and prediction-entropy proxies recover only $+0.06$ and $+0.03$ of that $+0.14$.

\begin{table*}[!t]
\centering
\caption{Analysis of Heterogeneity-Guided Stage-Adaptive Fusion on ISPRS Vaihingen}
\label{tab:haf}
\renewcommand{\arraystretch}{1.16}
\setlength{\tabcolsep}{5.0pt}
\footnotesize
\arrayrulecolor{rsrule}
\setlength{\tabcolsep}{13.0pt}
\begin{tabularx}{\linewidth}{l L cccc}
\tabtoprule
\rowcolor{rshead}
 & \textbf{Setting} & \textbf{Pixel-wise} & \textbf{Extra cue} & \textbf{mIoU} & \textbf{$\Delta$} \\
\tabheadrule
\multicolumn{6}{l}{\textit{Panel A: fusion-gate controls}}\\
H1 & Mean fusion & -- & -- & 84.28 & \down{0.20} \\
H2 & Global-pool gate & -- & GAP & 84.32 & \down{0.16} \\
H3 & Pixel gate, $B$ only & \cmark & -- & 84.34 & \down{0.14} \\
H4 & Pixel gate $+$ input gradient & \cmark & $\lvert\nabla I\rvert$ & \secv{84.40} & \down{0.08} \\
H5 & Pixel gate $+$ entropy & \cmark & $\mathbb{H}(p)$ & 84.37 & \down{0.11} \\
\rowcolor{rsmark}
H6 & \oursrow{HG-SAF ($B{+}\Hetero$)} & \cmark & $\Hetero$ & \bestv{84.48} & -- \\
\tabbotrule
\end{tabularx}
\setlength{\tabcolsep}{5.0pt}

\vspace{0.45em}
\begin{minipage}[t]{\dimexpr0.5\linewidth-0.6em\relax}
\begin{tabularx}{\linewidth}{L ccG}
\tabtoprule
\rowcolor{rshead}
\textbf{Stratum} & \textbf{UPerNet} & \textbf{$+$HG-SAF} & {\color{black}\textbf{$\Delta$}} \\
\tabheadrule
\multicolumn{4}{l}{\textit{Panel B: stratified IoU}}\\
Heterogeneity Q1 (lowest) & 86.12 & 86.28 & $+0.16$ \\
Heterogeneity Q2 & 84.05 & 84.41 & $+0.36$ \\
Heterogeneity Q3 & 81.74 & 82.38 & $+0.64$ \\
Heterogeneity Q4 (highest) & 76.84 & 78.41 & $+1.57$ \\
Small objects (area Q1) & 74.52 & 76.04 & $+1.52$ \\
Medium objects & 83.56 & 83.94 & $+0.38$ \\
Large objects & 87.12 & 87.31 & $+0.19$ \\
\tabbotrule
\end{tabularx}
\end{minipage}
\hfill
\begin{minipage}[t]{\dimexpr0.5\linewidth-0.6em\relax}
\renewcommand{\arraystretch}{1.4914}
\setlength{\tabcolsep}{11.0pt}
\begin{tabularx}{\linewidth}{L cc}
\tabtoprule
\rowcolor{rshead}
\textbf{Region} & \textbf{Shallow} & \textbf{Deep} \\
\tabheadrule
\multicolumn{3}{l}{\textit{Panel C: mean stage weights}}\\
Boundary & $0.37$ & $0.63$ \\
Small object & $0.38$ & $0.62$ \\
High $\Hetero$ & $0.37$ & $0.63$ \\
Low $\Hetero$ & $0.36$ & $0.64$ \\
Large interior & $0.35$ & $0.65$ \\
\tabbotrule
\end{tabularx}
\end{minipage}
\arrayrulecolor{black}
\tabnote{\emph{Panel~A} isolates the heterogeneity statistic $\Hetero$ from pixel-wise routing.  Every row replaces only the fusion stage and keeps the preparation blocks, FRA and CATP, so H1 coincides with the fusion-off row of \tabref{design}; H3--H6 additionally share the gate capacity of \equref{wsoftmax} and differ only in the extra cue.  \emph{Panel~B} reports IoU by heterogeneity quartile and by object size.  \emph{Panel~C} lists mean stage weights (shallow $=w_1{+}w_2$, deep $=w_3{+}w_4$), with Spearman $\rho(\Hetero,w_1{+}w_2)=0.10$ over the $113$ test patches; \figref{routing} shows the underlying maps and the pixel-wise distribution behind these two panels.  The deep pair holds the larger share in every stratum, so the effect is a shift in share rather than a change of the dominant stage.  As in \tabref{isprs}, Vaihingen scores include flip and rotation TTA, so the full model corresponds to $84.48\%$.  \bestsec{}  Ranking applies to the mIoU column of \emph{Panel~A}, where $\Delta$ is the drop with respect to H6; \emph{Panels~B} and~\emph{C} are paired measurements rather than a competition.}
\end{table*}

\begin{figure*}[!t]
\centering
\includegraphics[width=\linewidth]{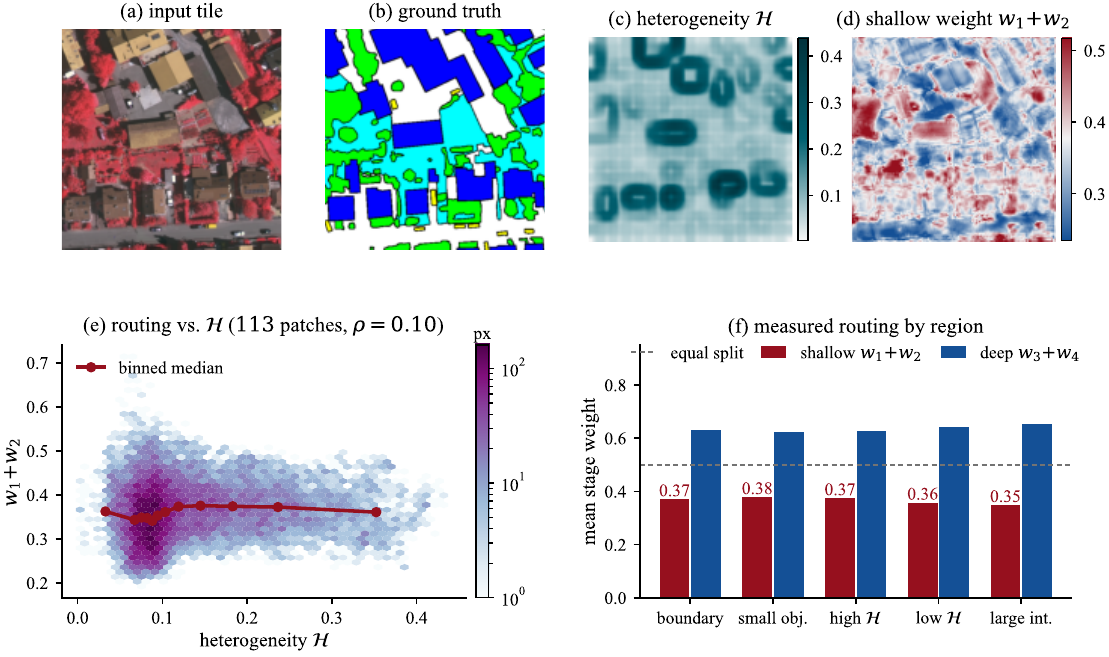}
\caption{Measured HG-SAF routing on ISPRS Vaihingen.  (a)~Input tile and (b)~ground truth.  (c)~The heterogeneity statistic of \equref{hetero} responds to object outlines and small structures.  (d)~Predicted shallow-stage weight $w_1{+}w_2$ of \equref{wsoftmax}, with the colour range set to the $1$--$99$ percentile of the tile so that the actual spatial variation is visible.  (e)~Pixel density of $w_1{+}w_2$ against $\Hetero$ over the $113$ test patches, with the binned median; the rank correlation over all pixels is $\rho{=}0.10$.  (f)~Mean stage weights per region, i.e.\ the quantities in Panel~C of \tabref{haf}.  The routing shift is consistent in sign across strata while deep stages keep the larger share everywhere.}
\label{fig:routing}
\end{figure*}

\begin{figure*}[t]
\centering
\includegraphics[width=0.96\linewidth]{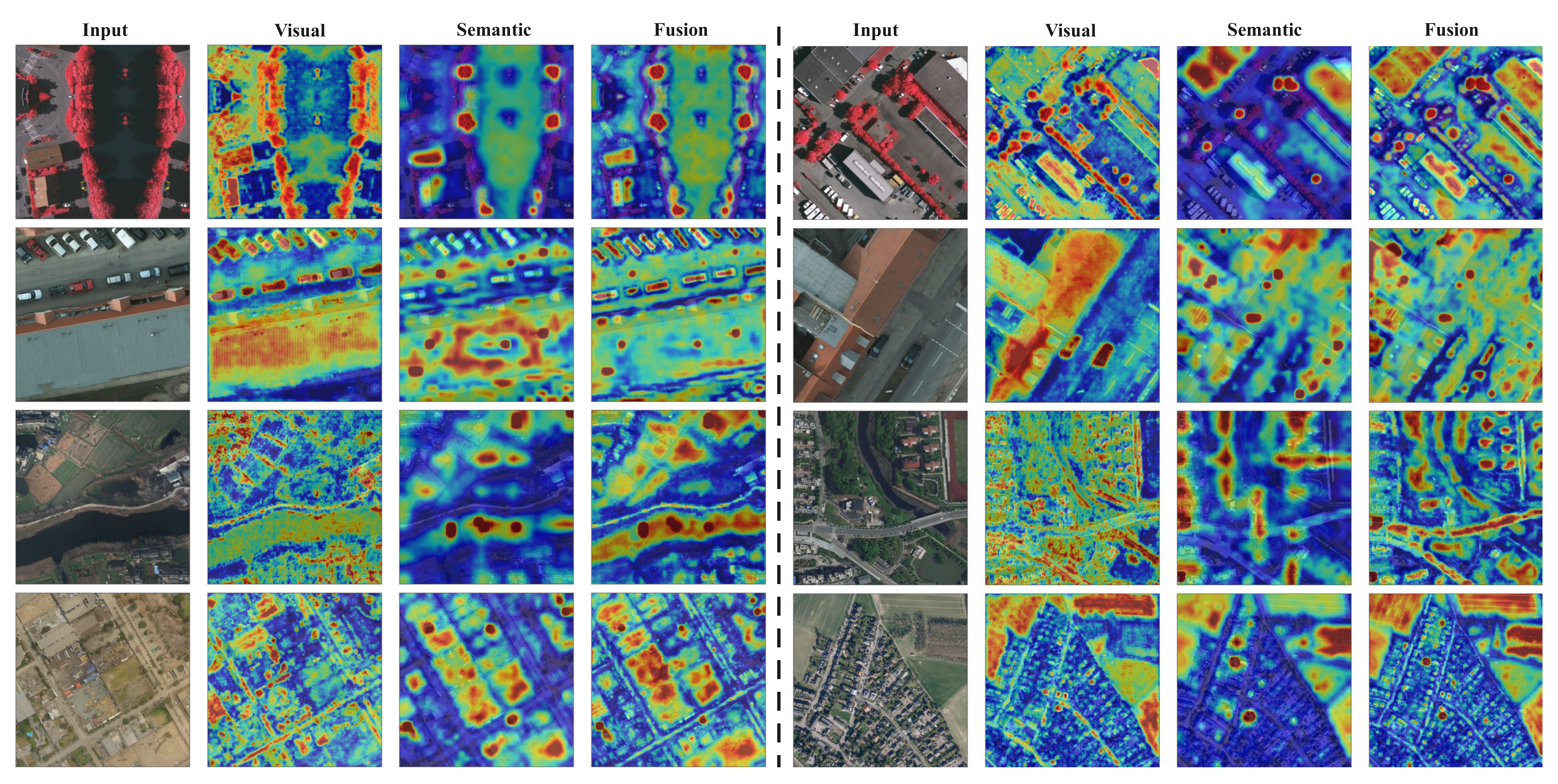}
\caption{Illustrative HG-SAF fusion on eight tiles (\textit{Input}, shallow visual cues, deep semantic context, fused map).  Quantitative routing statistics are in \tabref{haf}.}
\label{fig:feat-fuse}
\end{figure*}

Panel~B of \tabref{haf} shows that stratified IoU increases most in the highest quartile of \equref{hetero} ($+1.57$\,pp) and on small objects ($+1.52$\,pp), whereas large objects change by only $+0.19$\,pp.  Panel~C reports the corresponding routing statistics, and \figref{routing} shows them directly on the predicted maps.  The heterogeneity statistic tracks object outlines and small structures (\figref{routing}(c)), and the predicted weight field varies spatially at a comparable scale (\figref{routing}(d)).  The shift is consistent in sign across strata: shallow stages take a higher share on boundaries, small objects and high-$\Hetero$ pixels than in large interiors, the deep stages keep the larger share of the total weight everywhere, and the pixel-wise trend of $w_1{+}w_2$ against $\Hetero$ is positive (\figref{routing}(e),(f)).  HG-SAF therefore redistributes stage emphasis toward detail exactly in the strata where Panel~B reports the largest gains, while the deep semantic context of the pretrained pyramid remains dominant.  \figref{feat-fuse} provides a qualitative counterpart on eight tiles.

The fused maps in \figref{feat-fuse} retain shallow edge responses in heterogeneous regions and deeper homogeneous responses in large interiors.  This visualization is the qualitative counterpart of the routing statistics in Panel~C.  The next subsection asks whether a subsequent residual correction can further improve structural detail.

\subsection{Frequency-Residual Adaptation}
\label{sec:c2}
\tabref{bfr} compares FRA with matched spatial and spectral alternatives at the same fused-feature location.  The goal is to separate three possible sources of gain: the use of a residual branch, the use of a frequency operator, and the bounded low-rank parameterization.

\begin{table}[!t]
\centering
\caption{Matched Frequency-Module Alternatives on ISPRS Vaihingen}
\label{tab:bfr}
\renewcommand{\arraystretch}{1.16}
\setlength{\tabcolsep}{3.9pt}
\scriptsize
\arrayrulecolor{rsrule}
\begin{tabularx}{\linewidth}{L cccccc}
\tabtoprule
\rowcolor{rshead}
\textbf{Variant} & \textbf{Par.} & \textbf{mIoU} & \textbf{BIoU} & \textbf{BF} & \textbf{Thin} & \textbf{Small} \\
\tabheadrule
None (spatial only)        & --  & \secv{84.31} & 71.42 & 74.86 & 68.26 & 76.61 \\
Spatial residual (matched) & 1.1 & 84.22 & 71.78 & 75.14 & 68.58 & 76.84 \\
Standard FFT filter        & 1.2 & 84.19 & 72.06 & 75.41 & 68.94 & 76.72 \\
FFC-like branch~\cite{ffc} & 3.4 & 84.08 & 71.64 & 75.02 & 68.41 & 76.38 \\
GFNet mixer~\cite{gfnet}   & 4.8 & 83.96 & 71.28 & 74.71 & 68.05 & 76.12 \\
GFNet $+$ zero-init res.   & 4.8 & 84.28 & \secv{72.54} & \secv{75.88} & \secv{69.42} & \secv{77.05} \\
Full-rank spectral mixer   & 5.6 & 83.65 & 70.84 & 74.18 & 67.52 & 75.66 \\
Unbounded gate             & 1.2 & 83.88 & 71.16 & 74.52 & 67.88 & 75.94 \\
\rowcolor{rsmark}
\oursrow{FRA}              & 1.2 & \bestv{84.48} & \bestv{73.58} & \bestv{76.91} & \bestv{70.84} & \bestv{77.86} \\
\tabbotrule
\end{tabularx}
\arrayrulecolor{black}
\tabnote{Every variant replaces FRA at the same fused-feature location, and Vaihingen scores include flip and rotation TTA as in \tabref{isprs}, so the first row coincides with the frequency-off row of \tabref{design}.  Par.: added parameters (M); BIoU: Boundary IoU; BF: boundary $F$-score; Thin: thin-structure IoU; Small: small-component IoU.  \bestsec{}  The parameter column reports cost and is not ranked.}
\end{table}

A parameter-matched spatial residual and a standard FFT filter change mIoU only slightly, whereas unrestricted full-rank mixing is the weakest variant.  A GFNet mixer with the same zero-init residual parameterization already recovers part of the gain and is the second-best variant on four of the five accuracy metrics, which supports residual parameterization itself; the additional bounded low-rank design of FRA further improves Boundary IoU and thin-structure IoU.  Inspecting the learned gate confirms that it operates well inside the envelope that \equref{fra-gate} imposes: averaged over the $113$ Vaihingen test patches and over all coefficients of the half spectrum, its distance from identity is $|g{-}1|=0.28$, that is, $56\%$ of the $\alpha{=}0.5$ bound, so the branch is neither idle nor saturated at its limit.  The constraints of \equrange{fra-fft}{fra-out} apply to every coefficient alike, so the margin in \tabref{bfr} follows from the bounded low-rank parameterization rather than from a preference for any part of the spectrum.  FRA is therefore effective as a conservative spectral residual, and not as a generic frequency-module replacement.

\subsection{Tri-Prior Decoding}
\label{sec:c3}
\tabref{sad} compares the structural terms of \equref{loss} with conventional auxiliary losses at matched loss weights, and reports pairwise confusion mass on the frozen relation set.  The question is whether the tri-prior combination improves on adding the same families of loss as independent multi-task terms.

\begin{table}[!t]
\centering
\caption{Confusion-Aware Tri-Prior Decoder Analysis}
\label{tab:sad}
\renewcommand{\arraystretch}{1.16}
\setlength{\tabcolsep}{8.0pt}
\scriptsize
\arrayrulecolor{rsrule}
\begin{tabularx}{\linewidth}{l L cc}
\tabtoprule
\rowcolor{rshead}
& \textbf{Variant} & \textbf{mIoU} & \textbf{$\Delta$} \\
\tabheadrule
\multicolumn{4}{l}{\textit{Panel A: matched auxiliary-loss controls (Vaihingen)}}\\
S1 & CE$+$FocalDice only & 84.26 & \down{0.22} \\
S2 & $+$ boundary loss~\cite{boundaryloss} & 84.28 & \down{0.20} \\
S3 & $+$ objectness loss & 84.29 & \down{0.19} \\
S4 & $+$ pixel contrast~\cite{wang2022contrastive} & 84.31 & \down{0.17} \\
S5 & all three conventional losses & \secv{84.32} & \down{0.16} \\
\rowcolor{rsmark}
S6 & \oursrow{CATP (feedback $+$ pair hinge)} & \bestv{84.48} & -- \\
\tabbotrule
\end{tabularx}

\vspace{0.45em}
\begin{tabularx}{\linewidth}{L ccc}
\tabtoprule
\rowcolor{rshead}
\textbf{Pre-declared pair} & \textbf{w/o CATP} & \textbf{CATP} & \textbf{Rel.} \\
\tabheadrule
\multicolumn{4}{l}{\textit{Panel B: pairwise confusion mass $E_{a,b}$}}\\
Vaih.\ Imp.\ Surf.\ $\leftrightarrow$ Building & $0.142$ & $0.108$ & $-24\%$ \\
Vaih.\ Low Veg.\ $\leftrightarrow$ Tree & $0.118$ & $0.091$ & $-23\%$ \\
Pots.\ Imp.\ Surf.\ $\leftrightarrow$ Building & $0.096$ & $0.074$ & $-23\%$ \\
LoveDA Background $\leftrightarrow$ Barren & $0.186$ & $0.141$ & $-24\%$ \\
OEM Rangeland $\leftrightarrow$ Agriculture & $0.094$ & $0.068$ & $-28\%$ \\
\tabbotrule
\end{tabularx}
\arrayrulecolor{black}
\tabnote{\emph{Panel~A} compares CATP with conventional auxiliary losses on ISPRS Vaihingen under matched loss weights and epochs, so row S1 coincides with the decoder-off row of \tabref{design}; Vaihingen scores include flip and rotation TTA, as in \tabref{isprs}.  \emph{Panel~B} reports the pairwise confusion mass $E_{a,b}=M_{ab}+M_{ba}$ on the training-only pair set; \emph{Rel.} is the relative reduction against the matched model without CATP.  OEM: OpenEarthMap.  \bestsec{}  Ranking applies to the mIoU column of \emph{Panel~A}, where $\downarrow$ is the drop with respect to S6; a lower confusion mass is better throughout \emph{Panel~B}.}
\end{table}

The evidence that separates CATP from a generic multi-task head is Panel~B: pairwise confusion mass decreases by $23$--$28\%$ on every pre-declared pair of \equref{proto} that the panel reports, the largest reduction being OpenEarthMap \textit{Rangeland}\,$\leftrightarrow$\,\textit{Agriculture}, which is exactly where the hinge applies pressure.  Aggregate mIoU agrees: adding boundary, objectness or pixel-contrast losses individually improves the CE/FocalDice reference of \equref{focaldice} by at most $0.05$\,pp, and the three of them together remain $0.16$\,pp below CATP, so the gain comes from how the structural signals are coupled and not from the additional supervision alone.  The next subsection tests whether the three stages remain useful when they are combined.

\subsection{Component Complementarity and Initialization}
\label{sec:factorial}
\tabref{design} trains all eight combinations of HG-SAF, FRA and CATP and compares conservative initialization with random and fully active starts.  Combining the two analyses in one table avoids separating architecture from optimization.

\begin{table}[!t]
\centering
\caption{Component Complementarity and Initialization Study}
\label{tab:design}
\renewcommand{\arraystretch}{1.16}
\setlength{\tabcolsep}{3.0pt}
\scriptsize
\arrayrulecolor{rsrule}
\begin{tabularx}{\linewidth}{CCC ccccc}
\tabtoprule
\rowcolor{rshead}
\textbf{HG-SAF} & \textbf{FRA} & \textbf{CATP} & \textbf{Vaih.} & \textbf{Pots.} & \textbf{LoveDA} & \textbf{OEM} & \textbf{Avg.} \\
\tabheadrule
\multicolumn{8}{l}{\textit{Panel A: factorial combinations}}\\
-- & -- & -- & 83.81 & 87.30 & 53.91 & 66.32 & 72.84 \\
\cmark & -- & -- & 84.04 & 87.65 & 54.46 & 66.85 & 73.25 \\
-- & \cmark & -- & 83.99 & 87.52 & 54.18 & 66.61 & 73.08 \\
-- & -- & \cmark & 84.19 & 87.78 & 54.58 & 66.94 & 73.37 \\
\cmark & \cmark & -- & 84.26 & 87.93 & 54.81 & 67.27 & 73.57 \\
\cmark & -- & \cmark & \secv{84.31} & \secv{88.01} & \secv{54.92} & \secv{67.38} & \secv{73.66} \\
-- & \cmark & \cmark & 84.28 & 87.94 & 54.71 & 67.12 & 73.51 \\
\rowcolor{rsmark}
\cmark & \cmark & \cmark & \bestv{84.48} & \bestv{88.20} & \bestv{55.17} & \bestv{67.70} & \bestv{73.89} \\
\tabbotrule
\end{tabularx}

\vspace{0.45em}
\setlength{\tabcolsep}{7.0pt}
\begin{tabularx}{\linewidth}{L cccc}
\tabtoprule
\rowcolor{rshead}
\textbf{Initialization} & \textbf{mIoU} & \textbf{std$_{10}$} & \textbf{Best ep.} & \textbf{Div.} \\
\tabheadrule
\multicolumn{5}{l}{\textit{Panel B: alternative initializations (Vaihingen)}}\\
\rowcolor{rsmark}
\oursrow{Reference (ours)} & \bestv{84.48} & 0.42 & 67 & $0/3$ \\
Small random & \secv{84.19} & 0.71 & 71 & $0/3$ \\
Standard random & 84.12 & 0.91 & 74 & $0/3$ \\
Fully active & 83.88 & 1.24 & 76 & $1/3$ \\
\tabbotrule
\end{tabularx}
\arrayrulecolor{black}
\tabnote{\emph{Panel~A} lists the $2^{3}$ independently trained combinations of HG-SAF, FRA, and CATP; its first row switches all three off, leaving mean fusion and a single semantic head on the prepared stages.  mIoU increases along every path of the lattice on all four datasets, and pairwise interactions on Vaihingen are $I_{\mathrm{HF}}{=}+0.04$, $I_{\mathrm{HC}}{=}-0.11$ and $I_{\mathrm{FC}}{=}-0.09$\,pp, i.e., the modules are complementary and largely additive.  \emph{Panel~B} reports initialization stability over three seeds, where std$_{10}$ is the training-loss standard deviation over the first $10$ epochs, \emph{Best ep.} is the best epoch, and \emph{Div.} counts diverged runs.  ISPRS columns include flip and rotation TTA as in \tabref{isprs}; LoveDA and OpenEarthMap (OEM) use single-scale inference, so \emph{Avg.} here mixes the two inference settings and is not the matched no-TTA mean of \tabref{controlled}; it is used only to compare rows within this table.  \bestsec{}  Only the mIoU column is ranked in \emph{Panel~B}, because the remaining three columns are stability diagnostics for which a lower value is better.}
\end{table}

Each module is beneficial on its own, every pair improves on both of its members, and the full combination is best on all four datasets.  Restoring the missing module to a pair adds $+0.20$ for HG-SAF, $+0.17$ for FRA and $+0.22$\,pp for CATP on Vaihingen, and the pairwise interactions ($I_{\mathrm{HF}}{=}+0.04$, $I_{\mathrm{HC}}{=}-0.11$, $I_{\mathrm{FC}}{=}-0.09$\,pp, each formed as $m_{11}-m_{10}-m_{01}+m_{00}$) stay well below the main effects, so the three stages are complementary and largely additive.  Reference initialization attains the highest mIoU, the lowest early-loss variance, and no diverged run.  Starting from a known reference behavior therefore stabilizes fine-tuning and also improves its endpoint.

\subsection{Fixed Preparation Blocks}
\label{sec:trunk}
\tabref{trunk} measures the fixed blocks of \secref{prep} on the reference decoder, with HG-SAF, FRA and CATP switched off.

\begin{table}[!t]
\centering
\caption{Individual Effect of the Fixed Preparation Blocks}
\label{tab:trunk}
\renewcommand{\arraystretch}{1.16}
\setlength{\tabcolsep}{10.0pt}
\footnotesize
\arrayrulecolor{rsrule}
\begin{tabularx}{\linewidth}{l L cG}
\tabtoprule
\rowcolor{rshead}
& \textbf{Preparation} & \textbf{mIoU} & {\color{black}\textbf{$\Delta$}} \\
\tabheadrule
T1 & none (reference decoder) & 83.57 & {\color{black}--} \\
T2 & SSDB only                & 83.66 & $+0.09$ \\
T3 & BCS-Mamba only           & \secv{83.71} & $+0.14$ \\
T4 & SOE only                 & 83.64 & $+0.07$ \\
\rowcolor{rsmark}
T5 & \oursrow{SSDB $+$ BCS-Mamba $+$ SOE} & \bestv{83.79} & $+0.22$ \\
\tabbotrule
\end{tabularx}
\arrayrulecolor{black}
\tabnote{ISPRS Vaihingen mIoU (\%), single-scale and no TTA, so row~T1 is exactly the Swin-B$+$UPerNet reference of \tabref{controlled} and the $+0.22$\,pp of row~T5 is the share of the no-TTA margin of \tabref{controlled} that the preparation blocks contribute.  Only the preparation blocks change; HG-SAF, FRA and CATP are absent in all five rows.  \bestsec{}}
\end{table}

SSDB, BCS-Mamba and SOE contribute $+0.09$, $+0.14$ and $+0.07$\,pp individually and $+0.22$\,pp jointly, so row~T5 is the preparation stage adopted inside \hafr{}.  Two consequences are recorded explicitly.  First, of the $+0.55$\,pp that \hafr{} gains over the reference decoder on Vaihingen in \tabref{controlled}, $+0.22$\,pp comes from this fixed stage and $+0.33$\,pp from the three refinement stages.  Second, all module analyses in \tabrange{haf}{design} keep the preparation blocks fixed, so those comparisons are unaffected by this share.

\subsection{Qualitative Comparison}
\label{sec:qual}
\begin{figure*}[t]
\centering
\includegraphics[width=0.89\linewidth]{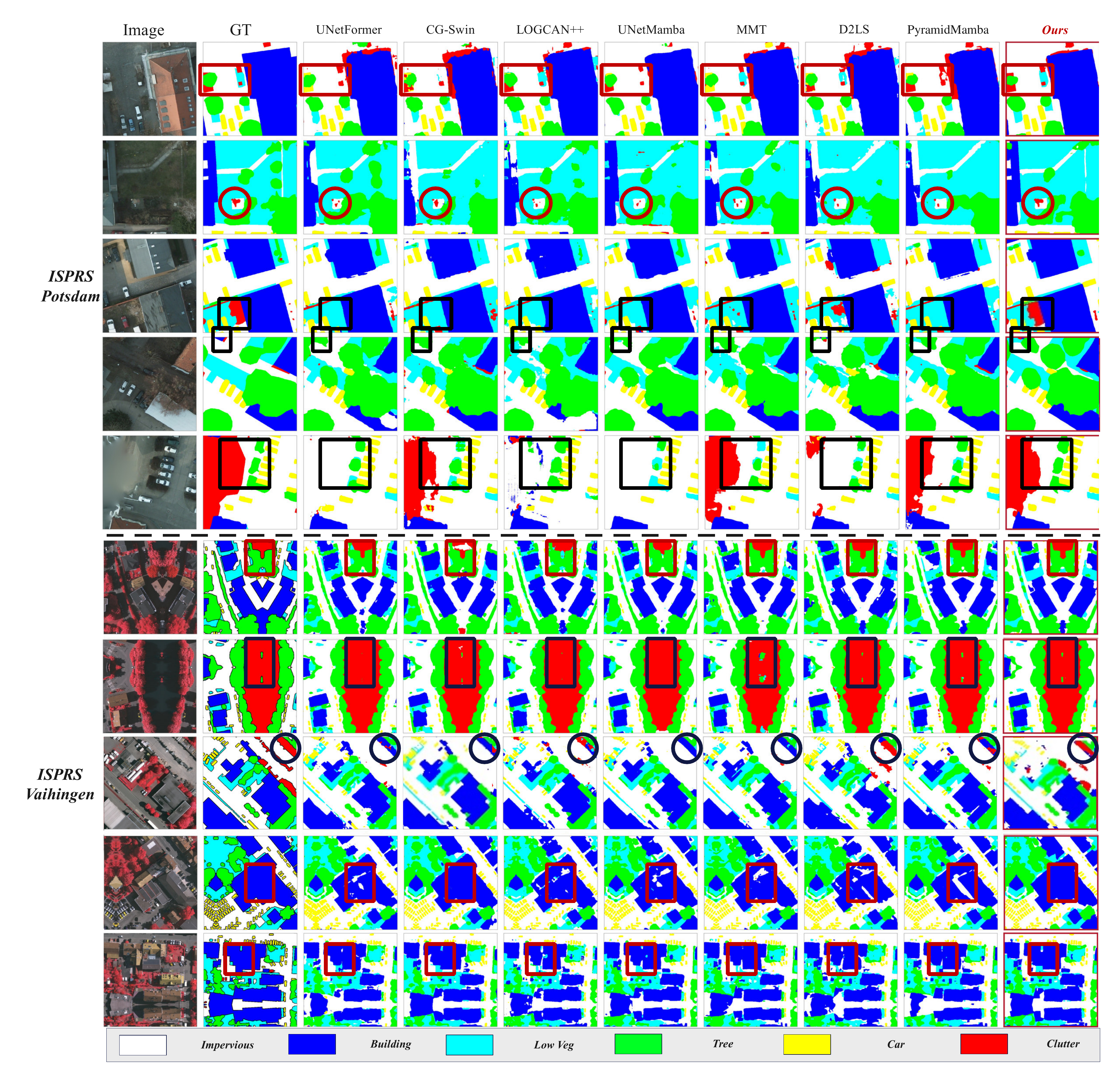}
\caption{Representative ISPRS examples.  Boxes mark vehicles and adjacent class boundaries discussed with the HG-SAF and CATP analyses.}
\label{fig:qual-isprs}
\end{figure*}

\begin{figure*}[t]
\centering
\includegraphics[width=0.93\linewidth]{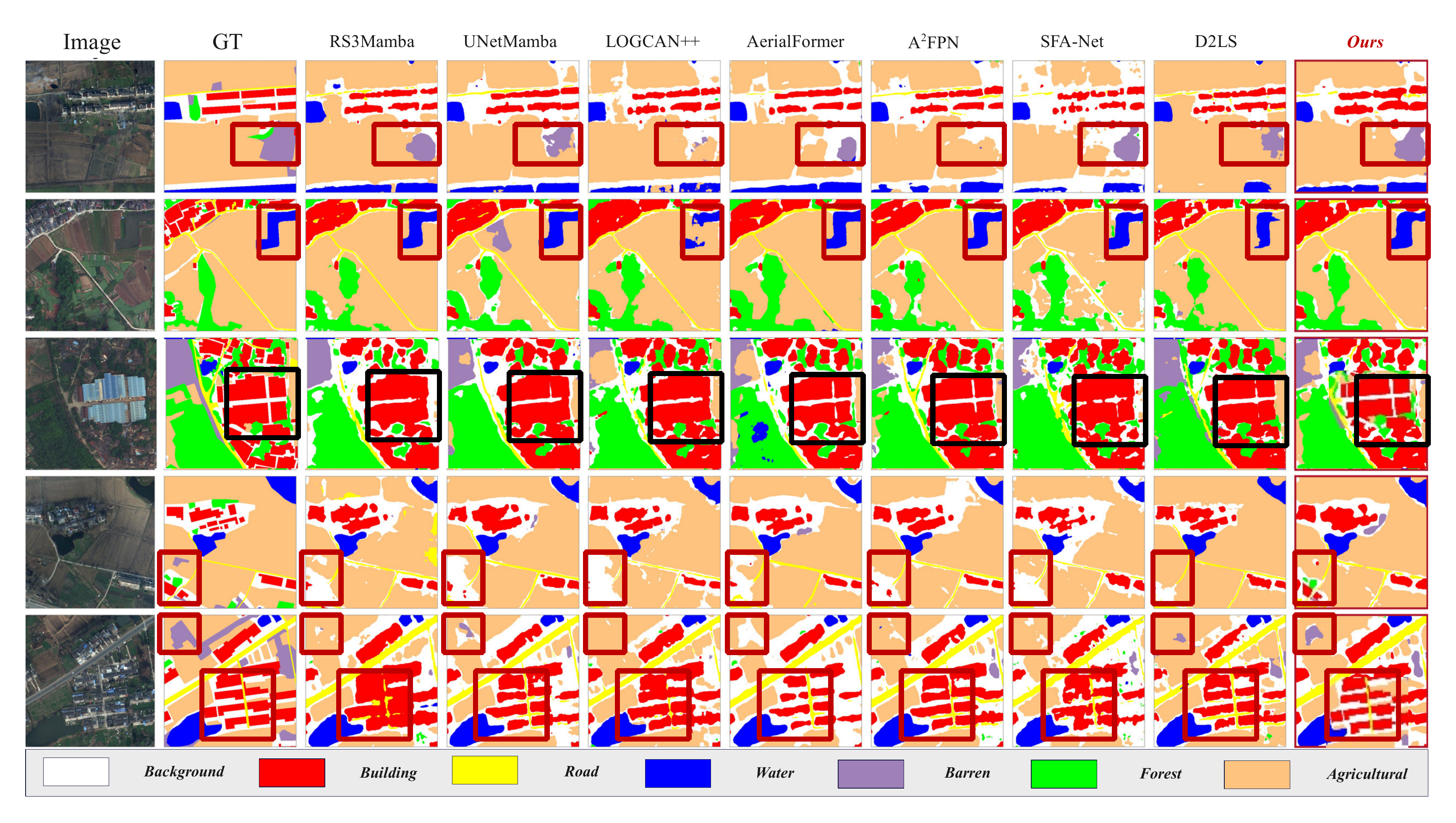}
\caption{Representative LoveDA examples.  Boxes mark thin rural roads and low-texture \textit{Barren} regions.}
\label{fig:qual-loveda}
\end{figure*}

\begin{figure*}[t]
\centering
\includegraphics[width=0.93\linewidth]{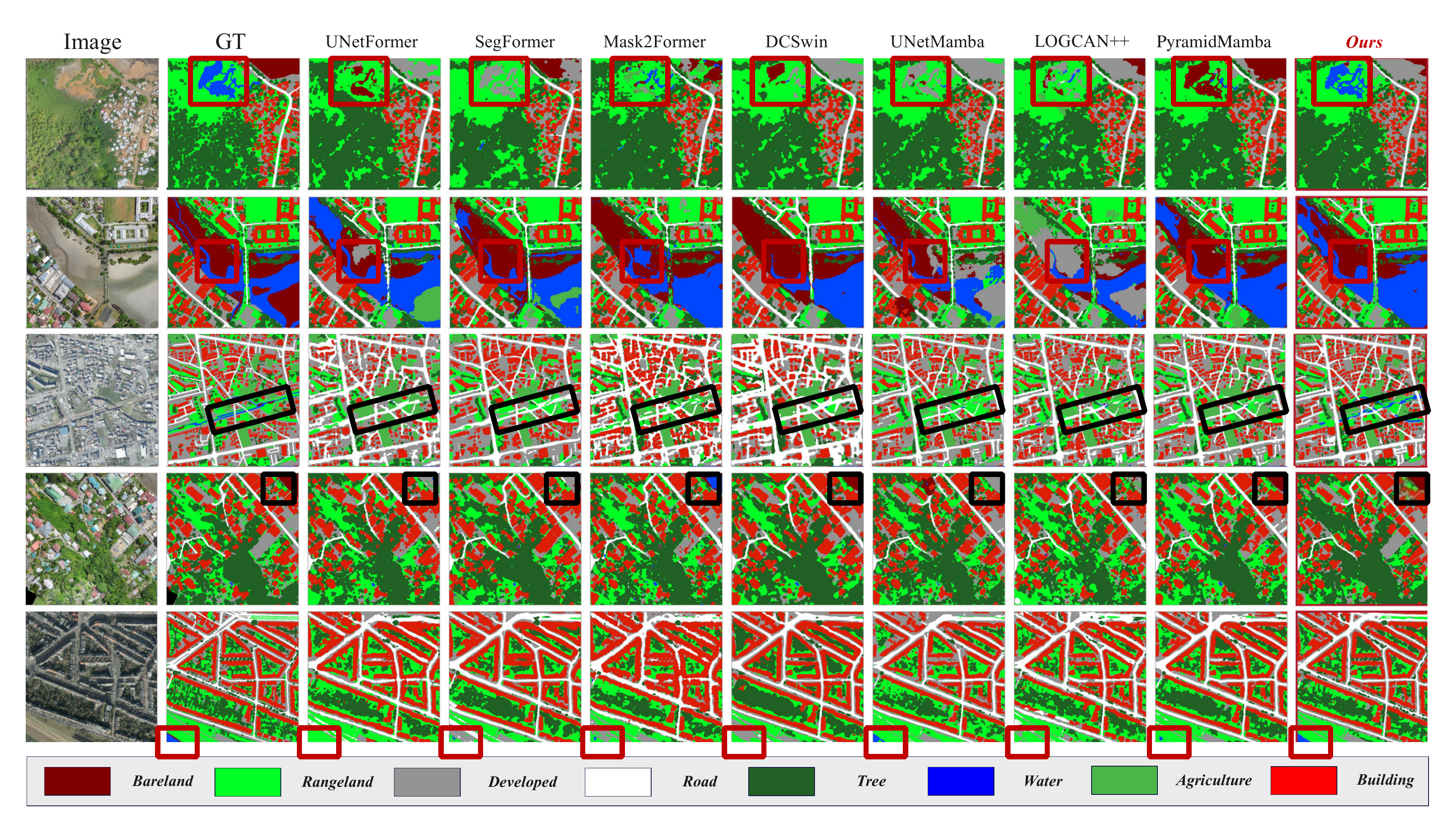}
\caption{Representative OpenEarthMap examples.  Boxes mark \textit{Rangeland}\,$\leftrightarrow$\,\textit{Agriculture} transitions and thin roads.}
\label{fig:qual-oem}
\end{figure*}

\figrange{qual-isprs}{qual-oem} close the analysis with representative predictions on the four benchmarks.  Every map shown, including those of the comparison methods, is produced by a model we trained ourselves and decoded with the same inference code; no prediction is reproduced from a published figure.  Each observation below belongs to a stratum that the analyses above already quantify, so the figures show where those measured effects become visible.  The same three tendencies recur across the four datasets.  Thin elongated structures stay connected over longer spans, adjacent instances of one class are merged into a single region less often, and the two classes of a pre-declared relation pair leak into each other less.  The regions we discuss are marked in each figure, and the errors that survive are collected in \secref{fail}.

The ISPRS examples in \figref{qual-isprs} concern vehicles adjacent to impervious surfaces and building outlines, which are the two regimes that the HG-SAF and CATP analyses measure separately.  Vehicles are the smallest annotated class in these benchmarks and fall almost entirely inside the small-object stratum of \tabref{haf}, where stratified IoU improves by $+1.52$\,pp in Panel~B and Panel~C shows the largest shift toward the shallow stages; in the marked parking rows the neighbouring vehicles stay separated rather than merging into one region.  Building outlines are the opposite case: long boundaries between two classes with overlapping colour statistics, i.e., the pair whose confusion mass CATP reduces in Panel~B of \tabref{sad}.  In the marked regions the predicted outline follows the annotated corner more closely.  Neither observation is specific to these tiles: both are consistent with the strata of \tabref{haf} and the pair panel of \tabref{sad}, and the boundary and small-component metrics of \tabref{bfr} improve alongside mIoU rather than in place of it.

The LoveDA examples in \figref{qual-loveda} highlight thin rural roads and low-texture \textit{Barren} patches.  Thin roads are exactly the regime that thin-structure IoU in \tabref{bfr} quantifies, and in the marked regions the road stays continuous where its surface is partly occluded instead of breaking into fragments.  \textit{Barren} is the harder case, because it is characterized less by a texture than by the absence of one, so a decoder that leans on local appearance tends to absorb it into \textit{Background}; that is the pair whose confusion mass CATP reduces in Panel~B of \tabref{sad}, and the marked patches keep one label over a larger area.  LoveDA also accounts for part of the dataset-level pattern in \tabref{controlled}: its tiles mix urban and rural domains, so a larger share of pixels falls in the upper heterogeneity quartiles, which is where Panel~B of \tabref{haf} reports the largest stratified gains.

The OpenEarthMap examples in \figref{qual-oem} show \textit{Rangeland}\,$\leftrightarrow$\,\textit{Agriculture} transitions and thin roads.  That pair is the pre-declared relation of this dataset and the one with the largest relative reduction of confusion mass in Panel~B of \tabref{sad}.  Its transition is gradual rather than sharp, so the decision is where to cut a broad gradient rather than where to find a visible edge, and in the marked regions the predicted boundary follows the field outline instead of oscillating inside the gradient.  OpenEarthMap carries the largest number of classes and the widest geographic spread of the four benchmarks, which is consistent with it also showing the largest matched-protocol margin in \tabref{controlled}.  Taken together, the four figures locate the measured gains in small objects, thin structures and pre-declared class pairs, and they do not show a uniform improvement over large homogeneous interiors, which is what Panel~B of \tabref{haf} already reports numerically.

The next section reports the compute cost of the full model and the remaining limitations.

\section{Efficiency and Limitations}
\label{sec:discussion}

\subsection{Efficiency}
\label{sec:eff}
Accuracy gains are useful only if their cost is stated under the same protocol as the controlled comparison.  \tabref{efficiency} reports parameters, GFLOPs, peak memory, latency and FPS for the re-implemented Swin-B family.  Published mixed-protocol models are omitted from that table.

\begin{table}[!t]
\centering
\caption{Efficiency of Five Re-Implemented Swin-B Decoders}
\label{tab:efficiency}
\renewcommand{\arraystretch}{1.16}
\setlength{\tabcolsep}{4.2pt}
\scriptsize
\arrayrulecolor{rsrule}
\begin{tabularx}{\linewidth}{L cccccc}
\tabtoprule
\rowcolor{rshead}
\textbf{Method} & \textbf{Par.} & \textbf{GFLOPs} & \textbf{Mem.} & \textbf{Lat.} & \textbf{FPS} & \textbf{mIoU} \\
\tabheadrule
Swin-B$+$FPN & 90.1 & 102.6 & 1.72 & 18.9 & 52.9 & 72.09 \\
Swin-B$+$UPerNet & 89.3 & 118.4 & 1.84 & 21.4 & 46.7 & 72.49 \\
Swin-B$+$UNetFormer dec. & 92.6 & 124.1 & 1.91 & 22.6 & 44.2 & \secv{72.78} \\
Swin-B$+$FFT mixer & 93.8 & 129.7 & 1.98 & 23.5 & 42.6 & 72.56 \\
\rowcolor{rsmark}
\oursrow{\hafr{}} & 97.8 & 131.2 & 2.12 & 24.8 & 40.3 & \bestv{73.71} \\
\tabbotrule
\end{tabularx}
\arrayrulecolor{black}
\tabnote{Cost is measured on one RTX~4090 under the timing protocol of \secref{metrics}, with TTA off.  Par.: parameters (M); Mem.: peak memory (GB); Lat.: latency (ms); mIoU is the four-dataset mean of \tabref{controlled} without TTA.  \bestsec{}  Only mIoU is ranked, because \hafr{} is the most expensive entry in every cost column.}
\end{table}

\begin{figure*}[!t]
\centering
\includegraphics[width=\linewidth]{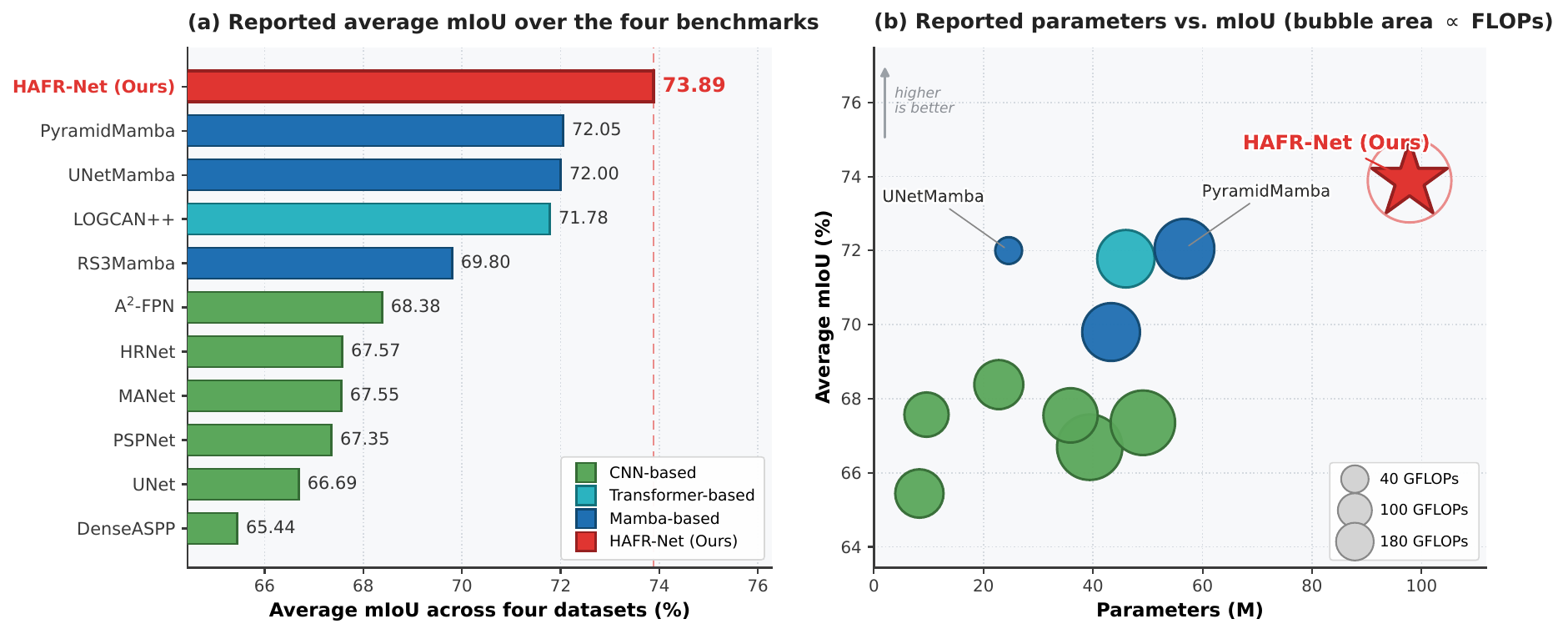}
\caption{Published accuracy and cost, shown as context.  Panel~(a) gives the four-dataset mean mIoU of the methods present in all three contextual tables; for \hafr{} this is the $73.89\%$ mean, which uses TTA on the two ISPRS sets.  Panel~(b) plots the reported parameter count against that mean, with bubble area proportional to the reported GFLOPs.  The entries come from different backbones and inference settings, so no frontier or optimality is drawn: the matched comparison is \tabref{controlled} and its measured cost \tabref{efficiency}.}
\label{fig:efficiency}
\end{figure*}

\hafr{} adds $8.5$\,M parameters over Swin-B$+$UPerNet.  The FP32 FFT of FRA costs $2.9$\,ms per $512{\times}512$ tile, i.e., about $12\%$ of the $24.8$\,ms forward pass of \tabref{efficiency}.  The accuracy gain in \tabref{controlled} is therefore obtained at a moderate, fully specified compute overhead.  \figref{efficiency} places the same mean next to published literature values.

Stating the same cost as a ratio against the reference makes the trade-off easier to compare with other decoders.  Relative to Swin-B$+$UPerNet, \hafr{} adds $9.5\%$ parameters, $10.8\%$ GFLOPs and $15.2\%$ peak memory, and it lowers throughput from $46.7$ to $40.3$\,FPS, i.e., by $13.7\%$, in exchange for $+1.22$\,pp of four-dataset mean mIoU without TTA.  Latency grows somewhat faster than GFLOPs ($15.9\%$ against $10.8\%$), which is consistent with the FP32 FFT of FRA being bound by memory traffic rather than by arithmetic; the ordering of the five methods in \tabref{efficiency} is the same under either measure.  We report FPS at batch size $1$, the setting in which that overhead is least amortized.  At $512{\times}512$ the full model still needs $24.8$\,ms per tile on one RTX~4090, so tiled inference over a large scene stays bounded by tiling and mosaicking rather than by the decoder.

Within \tabref{efficiency} the closest cost neighbour is the matched FFT mixer at $23.5$\,ms and $93.8$\,M parameters: \hafr{} spends $1.3$\,ms and $4.0$\,M more and returns $+1.15$\,pp of four-dataset mean mIoU.  The next subsection lists the cases that this cost does not resolve.

\subsection{Limitations}
\label{sec:fail}
Four limitations remain.  First, $\Hetero$ is a fixed functional of the learned feature $\tilde F_4$, namely the mean per-channel local standard deviation of \equref{hetero}.  It adds no parameters of its own, but it inherits whatever $\tilde F_4$ encodes, so a semantically homogeneous yet feature-variable region such as tree canopy can produce a boundary-like response.  The measured association between $\Hetero$ and the shallow-stage weight is correspondingly weak (\figref{routing}(e)), and a learned heterogeneity measure is left to future work.  Second, the spectral branches of SSDB and FRA evaluate their FFTs in FP32 for numerical stability, which is the latency reported in \secref{eff}.  A half-precision or windowed spatial approximation would remove that cost, but we have not verified that it preserves the boundary metrics of \tabref{bfr}, so we report the conservative variant.  Third, the prototype relation set depends on a training-only pilot split and does not adapt online, so a pair that becomes confusable only in a deployment domain receives no hinge; the number of candidate pairs also grows quadratically with the number of classes, which is why the set is declared in advance rather than searched.  Fourth, all experiments use RGB or NIR-RG VHR imagery and have not been extended systematically to SAR or hyperspectral data, where the spectral assumptions behind SSDB and FRA would have to be re-examined rather than transferred.  Residual errors that are not captured by hierarchical fusion, bounded refinement, or structural regularization---including cast shadows, mixed vegetation, and the catch-all \textit{Clutter} class---also remain difficult, and they dominate the marked regions of \figrange{qual-isprs}{qual-oem} that \hafr{} does not resolve.

\section{Conclusion}
\label{sec:conclusion}
This work studies VHR remote sensing image segmentation from the perspective of how pretrained hierarchical representations are decoded and refined.  Rather than adding unconstrained decoder capacity, we propose \hafr{}, which organizes the capacity it does add as a progressive refinement of the encoder hierarchy in three stages.  HG-SAF performs heterogeneity-conditioned pixel-wise fusion of multi-stage features, FRA introduces a bounded residual frequency correction inside a channel bottleneck, and CATP regularizes the final prediction with structural and class-relation cues.  Controlled experiments across four VHR benchmarks show gains of $+0.33$ to $+1.52$\,pp mIoU over the strongest matched baseline at a moderate and fully reported compute overhead, while module-specific analyses demonstrate spatially varying stage routing, improved boundary and thin-structure prediction, and reduced structural and class-confusion errors.  Adaptive and conservative refinement of strong pretrained representations is therefore an effective design principle for accurate VHR segmentation.  Future work will investigate learned heterogeneity measures, adaptive relation discovery, and other remote sensing modalities.

\section*{Acknowledgment}
The authors would like to thank the ISPRS, LoveDA, and OpenEarthMap organizers for releasing their benchmark datasets.

\bibliographystyle{IEEEtran}
\bibliography{references}

\end{document}